\documentclass[a4paper,11pt]{article}
\usepackage[thinlines]{easytable}
\usepackage{tikz}
\usepackage{float}
\usetikzlibrary{arrows, positioning, matrix, calc}
\usepackage{graphicx}
\usepackage{amsmath, amssymb, hyperref, graphicx, geometry, booktabs, array, caption, setspace}
\usepackage[square,sort,comma,numbers]{natbib}
\usepackage{lipsum}
\title{
    \hrule height 2pt
    \vspace{10pt}
    Optimal Scaling Laws for Efficiency Gains in a Theoretical Transformer-Augmented Sectional MoE Framework
    \vspace{10pt}
    \hrule height 1pt
}
\author{Soham Sane \\ \small Collins Aerospace }
\date{March $26^{th}, 2025$}

\begin{document}

\maketitle
\hrule height 2pt

\begin{abstract}
This paper introduces a theoretical framework for a Transformer-augmented, sectional Mixture-of-Experts (MoE) architecture that aims to enhance computational efficiency while preserving model scalability. Unlike conventional MoE models, which route entire token embeddings to selected experts, our approach partitions the embedding dimension itself—assigning segments of each token's representation to dedicated experts. To combat losses in token representation, we utilize a pre-expert transformer layer to recompute attention across tokens and reduce the sequence length dimensionality. We extend our theory by deriving optimal scaling laws that a non-linear relationship between the number of experts and factors such as model dimensionality, sequence length, and system overhead. These formulations yield closed-form and numerically-solvable expressions for identifying the optimal expert count under given architectural and hardware constraints. As a result, our framework not only provides theoretical bounds for computing efficiency with varying frameworks but also guides practical design choices for scaling large models effectively. While empirical validation is pending, we present a comprehensive experimental road map to evaluate the framework’s efficiency, scalability, and practicality in future work.
\end{abstract}

\newpage

\tableofcontents

\newpage

\section{Introduction}

Large-scale language models based on the Transformer architecture \cite{vaswani2017attention} have demonstrated remarkable success across a wide range of natural language processing tasks. As these models continue to scale in size and complexity, there is a growing need for architectures that can efficiently leverage computational resources while maintaining or improving performance. The Mixture-of-Experts (MoE) framework \cite{shazeer2017outrageously, fedus2022switch} has emerged as a promising approach to address these scaling challenges, offering a pathway to significantly increase model capacity without a proportional increase in computational demand.

\begin{figure}[H]
    \centering
    \includegraphics[width=0.5\linewidth]{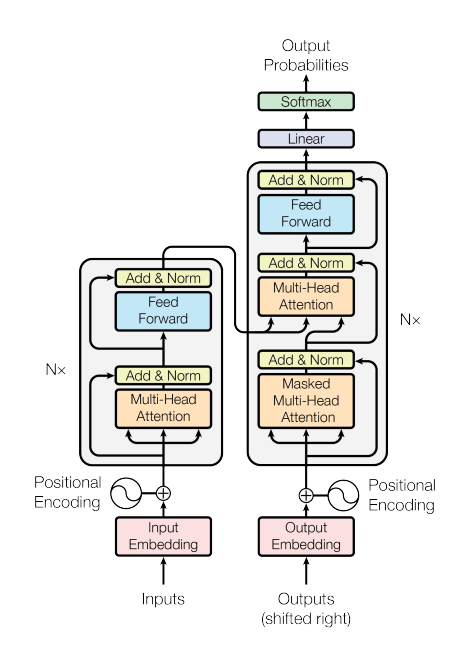}
    \caption{Transformer Architecture \cite{vaswani2017attention}}
    \label{fig:transformer}
\end{figure}

\subsection{Background and Related Work}

\subsubsection{Transformer Architecture}

The Transformer architecture, illustrated in Figure \ref{fig:transformer}, revolutionized sequence modeling by replacing recurrent neural networks with an attention-based mechanism. The fundamental components of the Transformer include:

\begin{itemize}
    \item \textbf{Multi-Head Attention}: This mechanism allows the model to attend to different positions in the input sequence simultaneously, capturing various types of dependencies. For each attention head, three projections—Query (Q), Key (K), and Value (V)—are computed from the input embeddings through linear transformations, followed by a scaled dot-product attention operation.
    
    \item \textbf{Position-wise Feed-Forward Networks (FFN)}: These are applied independently to each position in the sequence and consist of two linear transformations with a non-linear activation function in between.
    
    \item \textbf{Layer Normalization and Residual Connections}: These components stabilize training and facilitate gradient flow through the network.
\end{itemize}

The computational complexity of Transformer models scales quadratically with sequence length due to the attention mechanism, and linearly with model dimension. This has motivated extensive research into efficient variants that can maintain performance while reducing computational requirements \cite{tay2022efficient}.

\subsubsection{Mixture-of-Experts (MoE)}

The MoE framework \cite{jacobs1991adaptive, jordan1994hierarchical} decomposes complex tasks by employing specialized sub-networks (experts) controlled by a routing mechanism. In the context of Transformers, MoE layers typically replace the need for larger sequential transformer blocks by rather using smaller parallelized transformer blocks, allowing for an increase in model capacity without a corresponding increase in computational cost during inference \cite{shazeer2017outrageously}.

Some example variants of MoE architectures include:

\begin{itemize}
    \item \textbf{Dense MoE}: All experts process all tokens, with the outputs weighted by the router's softmax distribution. While comprehensive, this approach does not reduce computational complexity.
    
    \item \textbf{Sparse MoE}: Only a subset of experts (typically determined by a top-$k$ selection) processes each token. This significantly reduces computation but introduces load balancing challenges \cite{lepikhin2021gshard}.
    
    \item \textbf{Hierarchical MoE}: Employs a tree-like structure of expert routing, potentially enabling more efficient navigation of the expert space \cite{zuo2022taming}.
    
    \item \textbf{Conditional Computation MoE}: Dynamically determines which experts to activate based on input conditions, allowing for adaptive computation \cite{bengio2013estimating}.
\end{itemize}

\begin{figure}[H]
    \centering
    \includegraphics[width=1\linewidth]{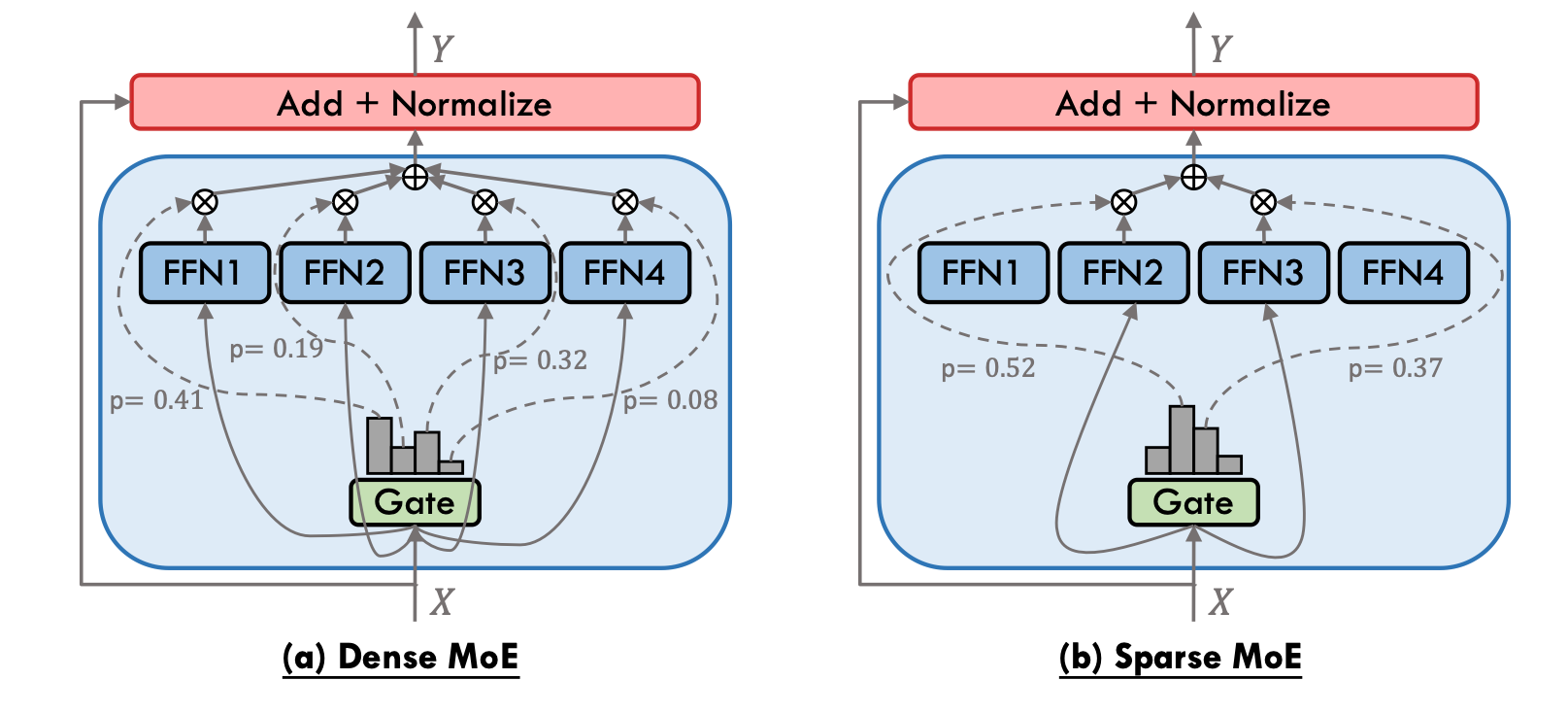}
    \caption{Example of a Dense and Sparse MoE Framework \cite{cai2024survey}}
    \label{fig:enter-label}
\end{figure}

\subsection{Motivations and Challenges}

The scaling of Transformer-based language models has been a major driving force behind recent advances in artificial intelligence. However, this scaling trajectory faces significant challenges:

\begin{itemize}
    \item \textbf{Computational Efficiency}: As model dimensions increase, the computational requirements grow quadratically, becoming prohibitively expensive for many practical applications. MoE models address this by activating only a subset of parameters for each input, effectively amortizing the computational cost across a larger parameter space.
    
    \item \textbf{Parameter Efficiency}: Traditional dense models require all parameters to be loaded into memory during inference. MoE architectures enable models with significantly larger parameter counts to operate within similar memory constraints by selectively activating experts.
    
    \item \textbf{Specialization}: MoE architectures allow for the development of specialized experts that can excel at different aspects of language understanding or generation, potentially leading to more nuanced and accurate models.
\end{itemize}
    
\noindent Despite these advantages, MoE models face several challenges:

\begin{itemize}
    \item \textbf{Load Balancing}: Ensuring uniform utilization of experts is non-trivial. Naive implementations often suffer from "expert collapse," where the router predominantly selects a small subset of experts \cite{lepikhin2021gshard, du2021glam}.
    
    \item \textbf{Routing Efficiency}: The process of determining which expert(s) should process each token introduces additional overhead, which can offset the computational savings if not carefully designed.
    
    \item \textbf{Training Stability}: MoE models can exhibit instability during training due to the discrete nature of expert selection and the co-adaptation between the router and experts \cite{yang2021stabilizing}.
    
    \item \textbf{Communication Overhead}: In distributed training scenarios, the communication patterns of MoE models are more complex than those of dense models, potentially introducing bottlenecks in the training process \cite{lewis2021base}.
\end{itemize}

\subsection{Contributions}

This paper introduces a novel sectionalized Mixture-of-Experts (MoE) framework that fundamentally reinterprets how input representations are distributed among experts. Our approach diverges from traditional MoE implementations by partitioning the embedding dimension rather than routing entire tokens to experts. In theory, this allows for a reduction of cross-domain expertise across the experts while reducing the computational cost from even a standard MoE framework. The key contributions of this work include:

\begin{enumerate}
    \item \textbf{Sectionalized Embedding Architecture}: We propose a framework that divides token embeddings along the feature dimension, with an additional attention layer, allowing each expert to process a slice of all tokens rather than complete embeddings of a subset of tokens while maintaining cross-token knowledge.
    
    \item \textbf{Theoretical Analysis}: We provide a rigorous mathematical derivation of the computational efficiency gains achieved by our approach, demonstrating significant reductions in the QKV computation cost with expert scaling.
    
    \item \textbf{Optimal Scaling Laws}: We derive closed-form expressions for the optimal number of experts as a function of model dimensions and system overhead, identifying the point at which further expert scaling yields diminishing returns.
    
    \item \textbf{Trade-off Analysis}: We analyze the inherent trade-offs between token specialization and cross-expert attention, showing how our approach balances computational efficiency with representational capacity.
\end{enumerate}

The remainder of this paper is organized as follows: Section 2 details the system architecture of traditional and sectionalized MoE frameworks, Section 3 derives the computational efficiency gains achieved by our approach, and Section 4 analyzes the optimal scaling laws and identifies the point of diminishing returns in expert scaling. Section 5 outlines experimentation strategies for future research and validation.

\newpage

\section{System Architecture of Traditional and Theoretical Sectionalized MoE Frameworks}

In this section, we compare the system architectures of a traditional Mixture-of-Experts (MoE) model and our proposed sectionalized MoE design. Although the derivations of the cost equations are presented in the subsequent section, here we focus on the flow of data, how the embeddings are processed, and the key differences between the two approaches.

\subsection{Traditional MoE Architecture}

In the traditional MoE framework, the input to the MoE layer is a tensor of shape \([EL, d_0]\), where \(L\) is the sequence length (number of tokens) multiplied by the number of experts \(L\) (assuming even and unique separation) and \(d_0\) is the full embedding dimension (see \cite{vaswani2017attention,shazeer2017outrageously}). The main steps are as follows:
\begin{enumerate}
    \item \textbf{Token Embedding:} Each token is embedded into a \(d_0\)-dimensional vector. The router (or gating network) receives the entire \([EL, d_0]\) tensor.
    \item \textbf{Gating and Routing:} The gating mechanism evaluates each token's embedding and routes the token to a subset of experts (e.g., the top-1 or top-2 experts). Thus, each expert receives the full \(d_0\)-dimensional embedding for only the tokens that it is selected to process.
    \item \textbf{Expert Computation:} Each expert performs its operations (such as the QKV projections) on the full token embedding. For example, if an expert processes a token, it handles an input of shape \([1, d_0]\) (or \([l, d_0]\) for a batch of \(k\) tokens). 
\end{enumerate}

\begin{figure}[H]
    \centering
    \begin{tikzpicture}[node distance=1cm, auto]
        \node[draw, rectangle, fill=blue!10] (input) {Input: \([EL, d_0]\)};
        \node[draw, rectangle, fill=green!10, below of=input] (router) {Router / Gating Network};
        \node[draw, rectangle, fill=yellow!20, below left of=router, node distance=3cm] (expert1) {Expert 1 \(\;[l_1, d_0]\)};
        \node[draw, rectangle, fill=yellow!20, below right of=router, node distance=3cm] (expert2) {Expert 2 \(\;[l_2, d_0]\)};
        \node[draw, rectangle, fill=red!10, below of=router, node distance=5cm] (aggregation) {Aggregation of Expert Outputs};
        
        \draw[->] (input) -- (router);
        \draw[->] (router) -- node[above left] {Token subset} (expert1);
        \draw[->] (router) -- node[above right] {Token subset} (expert2);
        \draw[->] (expert1) -- (aggregation);
        \draw[->] (expert2) -- (aggregation);
    \end{tikzpicture}
    \caption{Flow diagram for the traditional MoE framework. Each expert receives the full \(d_0\)-dimensional embedding for a subset of tokens.}
    \label{fig:traditional_moe}
\end{figure}
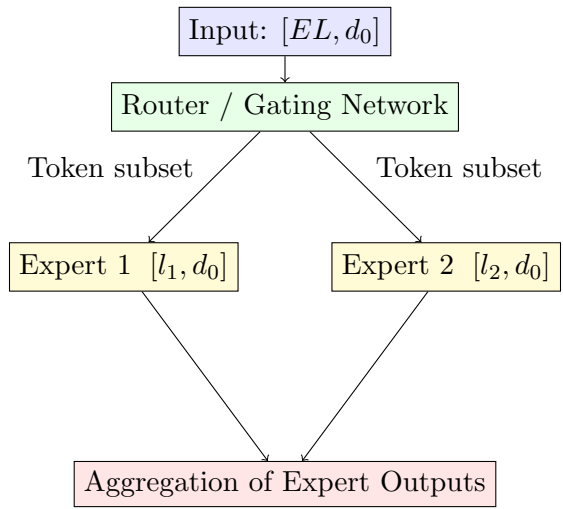

Figure~\ref{fig:traditional_moe} illustrates the flow process for a traditional MoE.

\subsection{Sectionalized MoE Architecture (Proposed)}

Our proposed design modifies the traditional framework by not only partitioning the embedding dimension among the experts but also by incorporating an additional pre-processing step before expert computation. The steps in this architecture are as follows:
\begin{enumerate}
    \item \textbf{Token Embedding:} Instead of the traditional \([L, d_0]\) split, we know perform a split along the \(L\)-dimensional vector, forming a tensor of shape \([L/E, d_{\text{slice}}]\) (inspired by efficient transformer designs.
    \item \textbf{Pre-Expert Processing:} However, due to the loss of the full embedding (as seen in the next step), we now include a transformer block to recover the lost dependencies with an attention mechanism and remake the \(L\)-dimensional vector back to the full length. The theory here is that the attention mechanism should be able to capture the lost dependencies caused by the dimension split as we are now also including a slice of each of the tokens across the entire input. Furthermore, we can also reduce the dimensionality of the the sequence length to achieve even greater computational gains.\\ 
    \\ Recent research on low-complexity attention offers promising avenues to further optimize these components. Notably, Katharopoulos et al. (2020) introduced a linear attention mechanism in Transformers are RNNs: Fast Autoregressive Transformers with Linear Attention" \cite{katharopoulos2020transformers}, which reduces the computational complexity of attention from \(O(L^2)\) to \(O(L)\). Similarly, Choromanski et al. (2021) in "Rethinking Attention with Performers" \cite{choromanski2021rethinking} provide another perspective on leveraging linear attention for efficient global dependency capture. Integrating these approaches within the pre-expert and post-expert attention blocks of the sectionalized MoE design could further reduce computational overhead, especially in scenarios involving long input sequences, while maintaining the model's ability to aggregate global context effectively.
    \item \textbf{Dimension Partitioning:} The refined embedding is then split into \(E\) equal slices along the embedding dimension. Each expert receives all reduced \(L/E\) tokens (post initial transformer block) but only a slice of the full embedding. Formally, each expert processes an input of shape \([L/E, d_{\text{slice}}]\) where 
    \[
        d_{\text{slice}} = \frac{d_0}{E}
    \]
    Note that \(L/E\) in this instance is representative of the dimension reduction by the pre-expert transformer layer and not representative of a split among the experts. This approach echoes techniques in dimensionality reduction and tensor decomposition. Foundational work by Kolda and Bader \cite{kolda2009tensor} provides a comprehensive theoretical basis for tensor decompositions, while Kim et al. \cite{kim2016hadamard} explore splitting and recombining feature dimensions via the Hadamard product in low-rank bilinear pooling. These methods inform our understanding of how to effectively partition high-dimensional embeddings. Utilizing the same theoretical foundation of feature recombination, we rather use the intuitive transformer layer to recover cross-token dependencies. It is important to note that this layer must be appropriately masked in decoder-only models (e.g., autoregressive language models) to preserve causality.\\ 
    \item \textbf{Expert Computation:} Each expert performs its operations (e.g., QKV projections) on its lower-dimensional slice. We can assume that this lower dimensional slice is the output of each of the transformer heads leading towards an expert out of our initial transformer block.
    \item \textbf{Aggregation and Integration:} The outputs from the experts, which represent different portions of the full embedding, are aggregated (for instance, via an additional transformer block) to reconstruct a complete token representation.
\end{enumerate}

\begin{figure}[H]
    \centering
    \begin{tikzpicture}[node distance=2cm, auto]
        \node[draw, rectangle, fill=blue!10] (input) {Input: \([EL, d_0]\)};
        \node[draw, rectangle, fill=red!10, below of=input] (transformer) {Transformer Reduction: \(EL \rightarrow L_{reduced}\) slices};
        \node[draw, rectangle, fill=green!10, below of=transformer] (split) {Embedding Split: \(d_{\text{slice}} \rightarrow d_0/E\)};
        
        \node[draw, rectangle, fill=orange!20, below left of=split, node distance=3cm] (attn1) {Attention \& FFN};
        \node[draw, rectangle, fill=yellow!20, below of=attn1] (expert1) {Expert 1 \([L_{\text{reduced}}, d_{\text{slice}}]\)};
        
        \node[draw, rectangle, fill=orange!20, below right of=split, node distance=3cm] (attn2) {Attention \& FFN};
        \node[draw, rectangle, fill=yellow!20, below of=attn2] (expert2) {Expert 2 \([L_{\text{reduced}}, d_{\text{slice}}]\)};
        
        \node[draw, rectangle, fill=red!10, below of=split, node distance=7cm] (agg) {Aggregation: Reconstruct Full Embedding};
        
        \draw[->] (input) -- (transformer);
        \draw[->] (transformer) -- (split);
        \draw[->] (split) -- node[above left] {Slice 1} (attn1);
        \draw[->] (split) -- node[above right] {Slice 2} (attn2);
        \draw[->] (attn1) -- (expert1);
        \draw[->] (attn2) -- (expert2);
        \draw[->] (expert1) -- (agg);
        \draw[->] (expert2) -- (agg);
    \end{tikzpicture}
    \caption{Flow diagram for the sectionalized MoE framework. The input embedding is first split into \(E\) slices. Each slice is processed by its own Attention \& FFN block before being fed to the corresponding expert. The outputs are then aggregated to reconstruct the full embedding.}
    \label{fig:sectionalized_moe}
\end{figure}
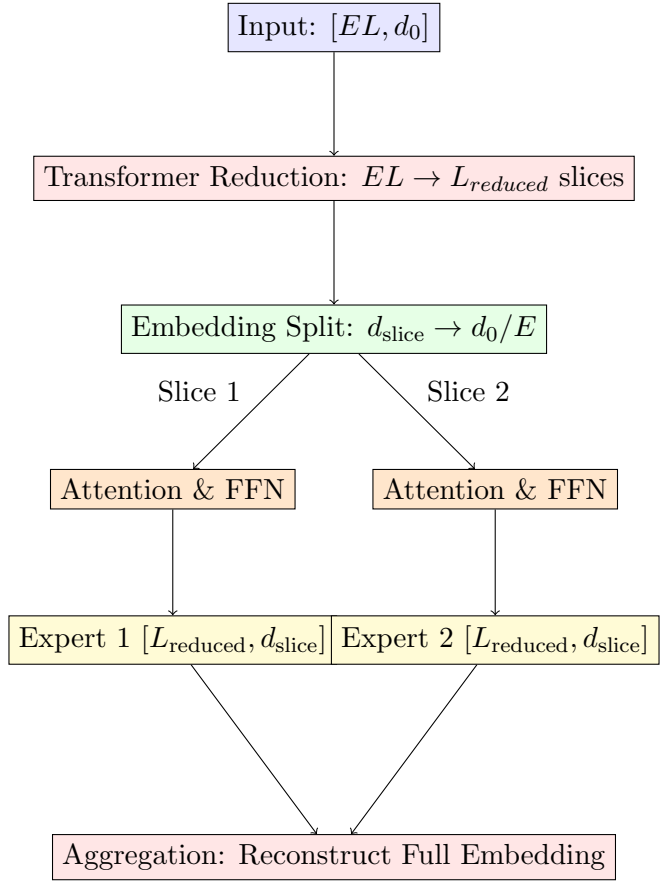

Figure~\ref{fig:sectionalized_moe} provides a flow diagram for the sectionalized MoE design.

\subsection{Implications and Trade-offs of the Sectionalized MoE Design}

Our proposed framework fundamentally alters the traditional MoE paradigm by splitting each token's embedding across experts. In a standard sparse MoE, the router directs tokens to experts that specialize in subsets of the token space, with each expert receiving the full token embedding.

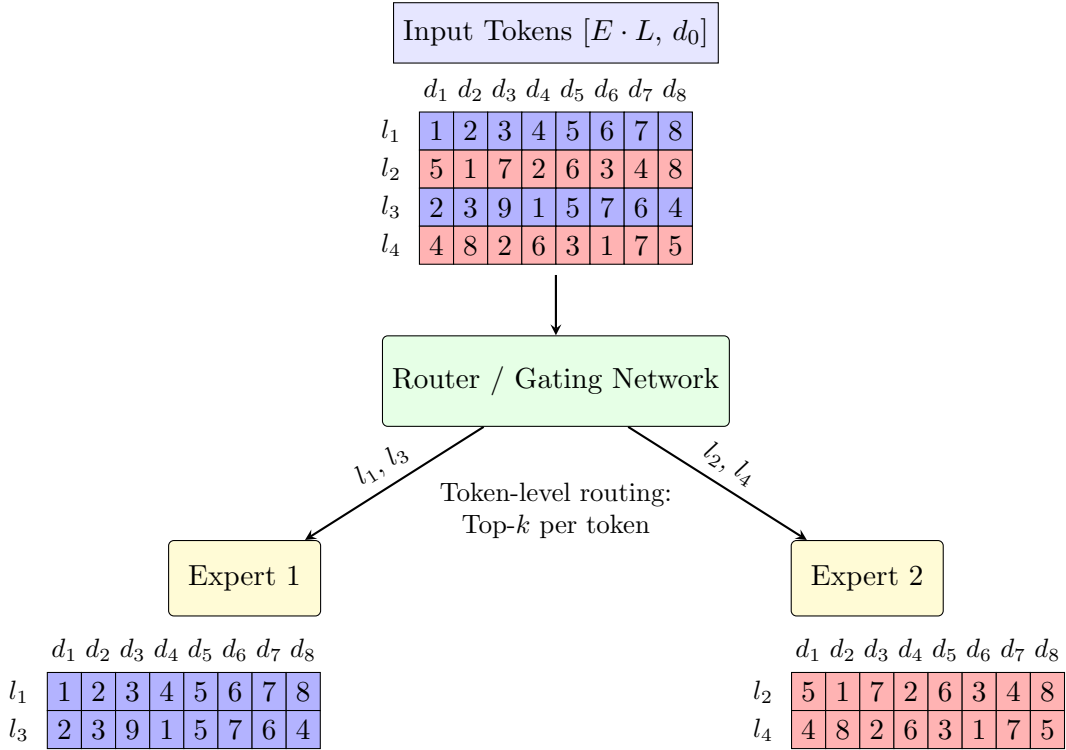
\begin{figure}[H]
    \centering
    \begin{tikzpicture}[
        node distance=1.5cm,
        box/.style={rectangle, draw, minimum width=3.5cm, minimum height=1.2cm, text centered, rounded corners=2pt},
        smallbox/.style={rectangle, draw, minimum width=1.5cm, minimum height=0.8cm, text centered},
        tinybox/.style={rectangle, draw, minimum width=0.5cm, minimum height=0.5cm, text centered, font=\tiny},
        arrow/.style={->, >=stealth, thick},
        tokenmatrix/.style={matrix of nodes, nodes={draw, minimum size=0.4cm}, column sep=-\pgflinewidth, row sep=-\pgflinewidth},
        tokenvalue/.style={rectangle, draw, minimum width=0.5cm, minimum height=0.3cm, text centered, font=\tiny},
        expert/.style={rectangle, draw, fill=yellow!20, minimum width=2cm, minimum height=1cm, text centered, rounded corners=2pt},
        section/.style={rectangle, draw=none, fill=white, text centered, font=\bfseries},
        data/.style={rectangle, draw, fill=blue!10, minimum width=3cm, minimum height=0.8cm, text centered}
    ]
    
    \node[section] (title) at (0, 0) {Traditional Sparse MoE};
    
    \node[data] (input) at (0, -1.2) {Input Tokens [$E \cdot L$, $d_0$]};
    
    \matrix[tokenmatrix, below=0.5cm of input] (m1) {
        |[fill=blue!30]| 1 & |[fill=blue!30]| 2 & |[fill=blue!30]| 3 & |[fill=blue!30]| 4 & |[fill=blue!30]| 5 & |[fill=blue!30]| 6 & |[fill=blue!30]| 7 & |[fill=blue!30]| 8 \\
        |[fill=red!30]| 5 & |[fill=red!30]| 1 & |[fill=red!30]| 7 & |[fill=red!30]| 2 & |[fill=red!30]| 6 & |[fill=red!30]| 3 & |[fill=red!30]| 4 & |[fill=red!30]| 8 \\
        |[fill=blue!30]| 2 & |[fill=blue!30]| 3 & |[fill=blue!30]| 9 & |[fill=blue!30]| 1 & |[fill=blue!30]| 5 & |[fill=blue!30]| 7 & |[fill=blue!30]| 6 & |[fill=blue!30]| 4 \\
        |[fill=red!30]| 4 & |[fill=red!30]| 8 & |[fill=red!30]| 2 & |[fill=red!30]| 6 & |[fill=red!30]| 3 & |[fill=red!30]| 1 & |[fill=red!30]| 7 & |[fill=red!30]| 5 \\
    };
    \node[left=0.1cm of m1-1-1, font=\small] {$l_1$};
    \node[left=0.1cm of m1-2-1, font=\small] {$l_2$};
    \node[left=0.1cm of m1-3-1, font=\small] {$l_3$};
    \node[left=0.1cm of m1-4-1, font=\small] {$l_4$};
    \node[above=0.01cm of m1-1-1, font=\small] {$d_1$};
    \node[above=0.01cm of m1-1-2, font=\small] {$d_2$};
    \node[above=0.01cm of m1-1-3, font=\small] {$d_3$};
    \node[above=0.01cm of m1-1-4, font=\small] {$d_4$};
    \node[above=0.01cm of m1-1-5, font=\small] {$d_5$};
    \node[above=0.01cm of m1-1-6, font=\small] {$d_6$};
    \node[above=0.01cm of m1-1-7, font=\small] {$d_7$};
    \node[above=0.01cm of m1-1-8, font=\small] {$d_8$};
    
    \node[box, fill=green!10, below=0.8cm of m1] (router) {Router / Gating Network};
    \draw[arrow] (m1) -- (router);
    
    \node[below=0.6cm of router, text width=6cm, align=center, font=\small] {Token-level routing:\\Top-$k$ per token};
    
    \node[expert, below left=1.5cm and 0.8cm of router] (expert1) {Expert 1};
    \draw[arrow] (router) -- node[above, sloped, font=\small] {$l_1$, $l_3$} (expert1);
    
    \node[expert, below right=1.5cm and 0.8cm of router] (expert2) {Expert 2};
    \draw[arrow] (router) -- node[above, sloped, font=\small] {$l_2$, $l_4$} (expert2);
    
    \matrix[tokenmatrix, below=0.6cm of expert1, xshift=-0.8cm] (m2) {
        |[fill=blue!30]| 1 & |[fill=blue!30]| 2 & |[fill=blue!30]| 3 & |[fill=blue!30]| 4 & |[fill=blue!30]| 5 & |[fill=blue!30]| 6 & |[fill=blue!30]| 7 & |[fill=blue!30]| 8 \\
        |[fill=blue!30]| 2 & |[fill=blue!30]| 3 & |[fill=blue!30]| 9 & |[fill=blue!30]| 1 & |[fill=blue!30]| 5 & |[fill=blue!30]| 7 & |[fill=blue!30]| 6 & |[fill=blue!30]| 4 \\
    };
    \node[left=0.1cm of m2-1-1, font=\small] {$l_1$};
    \node[left=0.1cm of m2-2-1, font=\small] {$l_3$};
    \node[above=0.01cm of m2-1-1, font=\small] {$d_1$};
    \node[above=0.01cm of m2-1-2, font=\small] {$d_2$};
    \node[above=0.01cm of m2-1-3, font=\small] {$d_3$};
    \node[above=0.01cm of m2-1-4, font=\small] {$d_4$};
    \node[above=0.01cm of m2-1-5, font=\small] {$d_5$};
    \node[above=0.01cm of m2-1-6, font=\small] {$d_6$};
    \node[above=0.01cm of m2-1-7, font=\small] {$d_7$};
    \node[above=0.01cm of m2-1-8, font=\small] {$d_8$};
    
    \matrix[tokenmatrix, below=0.6cm of expert2, xshift=0.8cm] (m3) {
        |[fill=red!30]| 5 & |[fill=red!30]| 1 & |[fill=red!30]| 7 & |[fill=red!30]| 2 & |[fill=red!30]| 6 & |[fill=red!30]| 3 & |[fill=red!30]| 4 & |[fill=red!30]| 8 \\
        |[fill=red!30]| 4 & |[fill=red!30]| 8 & |[fill=red!30]| 2 & |[fill=red!30]| 6 & |[fill=red!30]| 3 & |[fill=red!30]| 1 & |[fill=red!30]| 7 & |[fill=red!30]| 5 \\
    };
    \node[left=0.1cm of m3-1-1, font=\small] {$l_2$};
    \node[left=0.1cm of m3-2-1, font=\small] {$l_4$};
    \node[above=0.01cm of m3-1-1, font=\small] {$d_1$};
    \node[above=0.01cm of m3-1-2, font=\small] {$d_2$};
    \node[above=0.01cm of m3-1-3, font=\small] {$d_3$};
    \node[above=0.01cm of m3-1-4, font=\small] {$d_4$};
    \node[above=0.01cm of m3-1-5, font=\small] {$d_5$};
    \node[above=0.01cm of m3-1-6, font=\small] {$d_6$};
    \node[above=0.01cm of m3-1-7, font=\small] {$d_7$};
    \node[above=0.01cm of m3-1-8, font=\small] {$d_8$};
    
    \end{tikzpicture}
    \caption{Traditional MoE Framework Example. Please note that this assumes each expert receives unique tokens which is not always the case depending on the variation but it is drawn this way for demonstration purposes}
    \label{fig:traditional_moe_sparse}
\end{figure}

\vspace{10pt}
In contrast, our design partitions the embedding dimension among experts, so that each expert processes only a fraction (\(d_0/E\)) of each token's embedding with a reduced sequence dimensionality. This approach introduces several key implications and trade-offs:

\begin{itemize}
    \item \textbf{Loss of Token Specialization:}  
    By dividing the embedding across experts, we potentially lose the ability for any single expert to develop complete specialization on a specific subset of tokens. Instead of handling full-context representations individually, each expert only sees part of the token’s features.

    \item \textbf{Gain of Cross-Expert Attention:}  
    Despite the loss of token-level specialization, our framework introduces a powerful mechanism for aggregating information. With an additional attention layer and FFN applied prior to and after the splitting, the router is able to select a group of general experts that collectively capture global dependencies. This allows the system to maintain an understanding of the entire token context even though the individual experts operate on partial embeddings.

    \item \textbf{Cooperative Expert Processing:}  
    Rather than relying on a set of experts working in isolation on specialized tokens, our design encourages a cooperative process. The router selects a set of general experts, each processing different parts of the token representation, and then these partial outputs are aggregated to reconstruct a complete representation. In simpler terms, we are effectively assembling a group of experts to jointly answer the question, rather than having experts work individually on isolated aspects cf. \cite{fedus2022switch}).

    \item \textbf{Balancing Traditional and MoE Architectures:}  
    Our approach strikes a balance between a traditional transformer block, where each token's full representation is processed uniformly, and a sparse MoE (depicted in Figure~\ref{fig:traditional_moe_sparse}), where token specialization is emphasized. By using a robust router to select the best set of general experts and by integrating an attention mechanism across experts, we preserve the benefits of full-context processing while achieving significant computational savings by reducing the per-expert dimensionality.
\end{itemize}

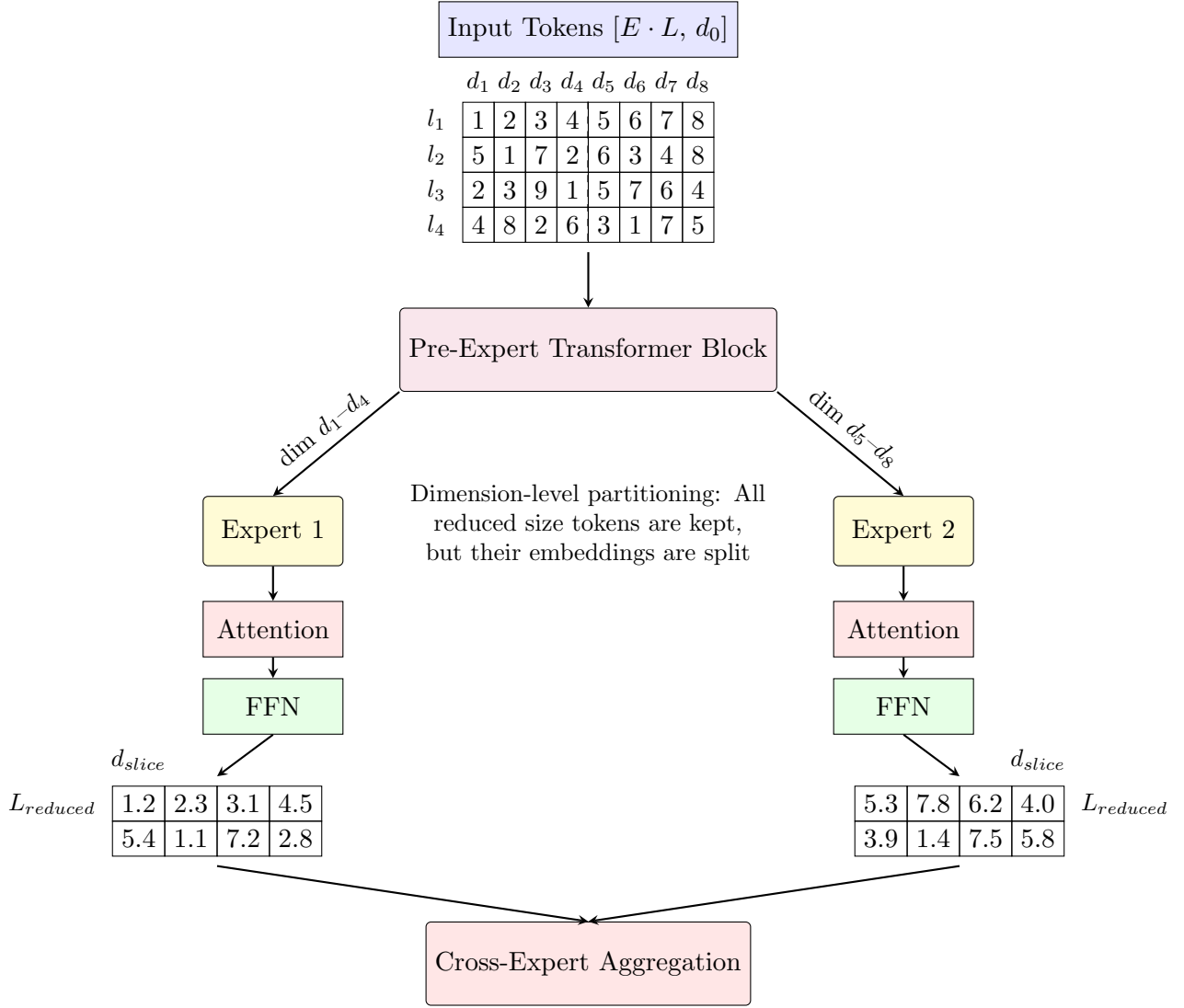
\begin{figure}[H]
    \centering
    \begin{tikzpicture}[
        node distance=1.5cm,
        box/.style={rectangle, draw, minimum width=3.5cm, minimum height=1.2cm, text centered, rounded corners=2pt},
        smallbox/.style={rectangle, draw, minimum width=1.5cm, minimum height=0.8cm, text centered},
        arrow/.style={->, >=stealth, thick},
        tokenmatrix/.style={matrix of nodes, nodes={draw, minimum size=0.4cm}, column sep=-\pgflinewidth, row sep=-\pgflinewidth},
        expert/.style={rectangle, draw, fill=yellow!20, minimum width=2cm, minimum height=1cm, text centered, rounded corners=2pt},
        section/.style={rectangle, draw=none, fill=white, text centered, font=\bfseries},
        data/.style={rectangle, draw, fill=blue!10, minimum width=3cm, minimum height=0.8cm, text centered},
        attention/.style={rectangle, draw, fill=red!10, minimum width=2cm, minimum height=0.8cm, text centered},
        ffn/.style={rectangle, draw, fill=green!10, minimum width=2cm, minimum height=0.8cm, text centered}
    ]
    
    \node[section] (title2) at (0,0) {Sectionalized MoE};
    
    \node[data] (input2) at (0,-1.2) {Input Tokens [$E \cdot L$, $d_0$]};
    
    \matrix[tokenmatrix, below=0.5cm of input2] (m4) {
        1 & 2 & 3 & 4 & 5 & 6 & 7 & 8 \\
        5 & 1 & 7 & 2 & 6 & 3 & 4 & 8 \\
        2 & 3 & 9 & 1 & 5 & 7 & 6 & 4 \\
        4 & 8 & 2 & 6 & 3 & 1 & 7 & 5 \\
    };
    \node[left=0.1cm of m4-1-1, font=\small] {$l_1$};
    \node[left=0.1cm of m4-2-1, font=\small] {$l_2$};
    \node[left=0.1cm of m4-3-1, font=\small] {$l_3$};
    \node[left=0.1cm of m4-4-1, font=\small] {$l_4$};
    \node[above=0.01cm of m4-1-1, font=\small] {$d_1$};
    \node[above=0.01cm of m4-1-2, font=\small] {$d_2$};
    \node[above=0.01cm of m4-1-3, font=\small] {$d_3$};
    \node[above=0.01cm of m4-1-4, font=\small] {$d_4$};
    \node[above=0.01cm of m4-1-5, font=\small] {$d_5$};
    \node[above=0.01cm of m4-1-6, font=\small] {$d_6$};
    \node[above=0.01cm of m4-1-7, font=\small] {$d_7$};
    \node[above=0.01cm of m4-1-8, font=\small] {$d_8$};
    
    \draw[dashed] ($(m4-1-5.north west)$) -- ($(m4-4-5.south west)$);
    
    \node[box, fill=purple!10, below=0.8cm of m4] (split) {Pre-Expert Transformer Block};
    \draw[arrow] (m4.south) -- (split.north);
    
    \node[below=1.2cm of split, text width=6cm, align=center, font=\small] {Dimension-level partitioning: All reduced size tokens are kept, but their embeddings are split};
    
    \node[expert, below left=1.5cm and 0.8cm of split] (expert1) {Expert 1};
    \draw[arrow] (split.south west) -- node[above, sloped, font=\small] {dim $d_1$--$d_4$} (expert1.north);
    
    \node[attention, below=0.5cm of expert1] (attention1) {Attention};
    \node[ffn, below=0.3cm of attention1] (ffn1) {FFN};
    \draw[arrow] (expert1.south) -- (attention1.north);
    \draw[arrow] (attention1.south) -- (ffn1.north);
    
    \node[expert, below right=1.5cm and 0.8cm of split] (expert2) {Expert 2};
    \draw[arrow] (split.south east) -- node[above, sloped, font=\small] {dim $d_5$--$d_8$} (expert2.north);
    
    \node[attention, below=0.5cm of expert2] (attention2) {Attention};
    \node[ffn, below=0.3cm of attention2] (ffn2) {FFN};
    \draw[arrow] (expert2.south) -- (attention2.north);
    \draw[arrow] (attention2.south) -- (ffn2.north);
    
    \matrix[tokenmatrix, below=0.6cm of ffn1, xshift=-0.8cm] (m5) {
        1.2 & 2.3 & 3.1 & 4.5 \\
        5.4 & 1.1 & 7.2 & 2.8 \\
    };
    \node[left=0.1cm of m5-1-1, font=\small] {$L_{reduced}$};
    \node[above=0.1cm of m5-1-1, font=\small] {$d_{slice}$};
    \draw[arrow] (ffn1.south) -- (m5.north);
    
    \matrix[tokenmatrix, below=0.6cm of ffn2, xshift=0.8cm] (m6) {
        5.3 & 7.8 & 6.2 & 4.0 \\
        3.9 & 1.4 & 7.5 & 5.8 \\
    };
    \node[right=0.1cm of m6-1-4, font=\small] {$L_{reduced}$};
    \node[above=0.1cm of m6-1-4, font=\small] {$d_{slice}$};
    \draw[arrow] (ffn2.south) -- (m6.north);
    
    \node[box, fill=red!10, below=0.8cm of $(m5.south)!0.5!(m6.south)$] (agg) {Cross-Expert Aggregation};
    \draw[arrow] (m5.south) -- (agg.north);
    \draw[arrow] (m6.south) -- (agg.north);
    
    \end{tikzpicture}
    \vspace{10pt}
    \caption{Sectionalized MoE Framework Example (Note: This does not include the router block as the routing can be specific to which MoE framework is followed)}
    \label{fig:sectionalized_moe_example}
\end{figure}

Overall, our design, as shown in Figure~\ref{fig:sectionalized_moe_example}, reinterprets the role of experts: instead of crafting a group of specialists, we create a team of general experts that collaborate to form a comprehensive solution, leveraging both the efficiency of reduced-dimensional computations and the power of global attention.

\newpage

\section{Deriving Computation Efficiency Gains}

In order to derive the computational efficiency that we will gain from this framework, we first define the costs associated with each computation within the current MoE framework and our implementation.

\subsection{Traditional MoE Approach}

\subsubsection{QKV Computation Cost for Traditional MoE \texorpdfstring{$(\mathcal{A})$}:}  This represents the cost of computing the queries (Q), keys (K), and values (V). For each token in the input sequence (of length \(L\)), we multiply the token’s embedding (of dimension \(d_0\)) by each of the three weight matrices \(W_q\), \(W_k\), and \(W_v\) (each of size \(d_0 \times d_0\)). Each multiplication requires \(d_0^2\) operations per token (as per standard matrix multiplication cost \cite{vaswani2017attention}), so for three matrices, the cost per token is \(3d_0^2\). Thus, over \(L\) tokens, the total cost is:
    \begin{equation}
        \mathcal{A}_{\text{trad}} = L \cdot 3 d_0^2
    \end{equation}
    
\subsubsection{Attention Score Calculation Cost for Traditional MoE \texorpdfstring{$(\mathcal{R})$}:} This accounts for the cost of the attention score computations, which consist of two main steps:
\begin{enumerate}
    \item Computing the dot product \(QK^T\). Given \(Q\) of size \([L, d_0]\) and \(K\) of size \([L, d_0]\) (transposed), this operation requires \(L^2 \cdot d_0\) operations.
    \item Multiplying the resulting attention scores with the value matrix \(V\) (of size \([L, d_0]\), which also requires \(L^2 \cdot d_0\) operations.
\end{enumerate}
Thus, the total cost is:
\begin{equation}
        \mathcal{R}_{\text{trad}} = L^2 d_0
    \end{equation}

\subsection{Proposed Sectionalized MoE Approach}

In our approach, the post-embedded output of size \( [LE, d_0] \) is first processed by a pre-expert transformer block before being split evenly along the embedding dimension among \( E \) experts. The pre-expert block receives the full token sequence of length \( {L}{E} \) of the router (with the full embedding dimension \( d_0 \)) and performs its own QKV and attention operations. The experts then each process an input slice of size \( \left[\frac{L}{E}, \frac{d_0}{E}\right] \). Note that \( L \) is used to signify an input sequence size for an individual expert and therefore we must multiply by \( E \) to calculate the full router input. This multiplication takes the assumption that experts have equal sizes in length such as those found in Dense MoE frameworks, but for the instances where that is not the case, the same equations can be modified to derive the reduction factors for that specific architecture.

\subsubsection{QKV Computation Cost for Sectionalized MoE:}

We consider two components in the QKV computation cost: a pre-expert transformer block and the expert blocks.

\paragraph{Pre-Expert Transformer Block:}  
The pre-expert block operates on \( L \cdot E\) tokens (as \( L\) is defined for a single expert in our calculations) with the full embedding \( d_0 \). Its QKV computation cost is:
\begin{equation}
    \mathcal{A}_{\text{pre}} = \ L \cdot E \cdot 3d_0^2
\end{equation}

\paragraph{Expert Blocks:}  
After the pre-expert block, the post-attention sequence and embedding is split into \( E \) slices respectively, where \( d_{\text{slice}} = \frac{d_0}{E} \) \( L_{\text{slice}} = \frac{L}{E} \) . For each expert, the QKV cost is:
\begin{equation}
    \mathcal{A}_{\text{expert}} = \frac{L}{E} \cdot 3\left(\frac{d_0}{E}\right)^2
\end{equation}
Summing over all experts:
\begin{equation}
    \mathcal{A}_{\text{experts}} = E \cdot \frac{L}{E} \cdot 3\left(\frac{d_0}{E}\right)^2 = L \cdot \frac{3d_0^2}{E^2}
\end{equation}

\noindent \textbf{Combined QKV Cost:}  
The total QKV computation cost is given by the sum of the pre-expert and expert components:
\begin{equation}
    \mathcal{A}_{\text{total}}^{\text{new}} = \mathcal{A}_{\text{pre}} + \mathcal{A}_{\text{experts}} = \ L \cdot E \cdot 3d_0^2 + L \cdot \frac{3d_0^2}{E^2} = \frac{3Ld_0^2(E^3 + 1)}{E^2}
\end{equation}

\noindent \textbf{Note on Attention Heads:}  
The above formulation assumes a uniform number of attention heads across all transformer blocks. In practice, however, it is possible that the pre-expert transformer block and the expert blocks use different numbers of attention heads—denoted \( H_{\text{pre}} \) and \( H_{\text{exp}} \), respectively. If we explicitly account for this, the QKV cost in each component changes due to the reduced per-head dimensionality.\\

For each attention head, the QKV computation involves projecting the input embedding of size \( d_0 \) to a subspace of dimension \( d_h = \frac{d_0}{H} \). This means the cost of one QKV projection per head is \( 3 d_0 \cdot d_h = 3 \frac{d_0^2}{H} \), and across all heads, the total QKV cost remains \( 3 d_0^2 \). However, if the number of heads differs between blocks, the costs no longer cancel cleanly in the reduction factor.\\

Specifically, the QKV cost for the pre-expert block becomes:
\begin{equation}
    \mathcal{A}_{\text{pre}} = {L \cdot E} \cdot H_{\text{pre}} \cdot 3 \left( \frac{d_0}{H_{\text{pre}}} \right)^2 = \frac{3EL d_0^2}{ H_{\text{pre}}}
\end{equation}

Similarly, for each of the \( E \) experts with \( H_{\text{exp}} \) heads and embedding dimension \( \frac{d_0}{E} \), the QKV cost per expert is:
\begin{equation}
    \mathcal{A}_{\text{expert}} = \frac{L}{E} \cdot H_{\text{exp}} \cdot 3 \left( \frac{d_0}{E H_{\text{exp}}} \right)^2 = \frac{3L d_0^2}{E^3 H_{\text{exp}}}
\end{equation}

Summing across all \( E \) experts yields:
\begin{equation}
    \mathcal{A}_{\text{experts}} = E \cdot \frac{3L d_0^2}{E^3 H_{\text{exp}}} = \frac{3L d_0^2}{E^2 H_{\text{exp}}}
\end{equation}

\noindent \textbf{Combined QKV Cost with Variable Head Counts:}
\begin{equation}
    \mathcal{A}_{\text{total}}^{\text{new}} = \frac{3EL d_0^2}{H_{\text{pre}}} + \frac{3L d_0^2}{E^3 H_{\text{exp}}}
\end{equation}

\subsubsection{Attention Score Calculation Cost for Sectionalized MoE:}

Again, we split the attention cost into two parts.

\paragraph{Pre-Expert Transformer Block:}  
The pre-expert block processes the sequence of \( {L}\cdot {E} \) tokens. Its attention score computation involves:
\begin{enumerate}
    \item Computing the dot product \( QK^T \) on matrices of size \( \left[{L}\cdot{E}, d_0\right] \), costing \( \left({L}{E}\right)^2 d_0 \) operations.
    \item Multiplying the resulting scores with the value matrix \( V \), also costing \( \left({L}{E}\right)^2 d_0 \) operations.
\end{enumerate}
Thus, for this additional computation, the cost is:
\begin{equation}
    \mathcal{R}_{\text{pre}} =  2\left({L}{E}\right)^2 d_0 = 2{EL^2 d_0}
\end{equation}

\paragraph{Expert Blocks:}  
Each expert computes attention on matrices of size \( \left[\frac{L}{E}, \frac{d_0}{E}\right] \), with a cost per expert of:
\begin{equation}
    \mathcal{R}_{\text{expert}} = 2\left(\frac{L}{E}\right)^2\left(\frac{d_0}{E}\right)
\end{equation}
Summing over all \( E \) experts yields:
\begin{equation}
    \mathcal{R}_{\text{experts}} = E \cdot 2\left(\frac{L}{E}\right)^2\left(\frac{d_0}{E}\right) = \frac{2L^2{d_0}}{E^2}
\end{equation}

\noindent \textbf{Combined Attention Cost:}  
The overall attention score computation cost is:
\begin{equation}
    \mathcal{R}_{\text{total}}^{\text{new}} = \mathcal{R}_{\text{pre}} + \mathcal{R}_{\text{experts}} = 2{EL^2 d_0} + \frac{2L^2{d_0}}{E^2}
\end{equation}

\noindent \textbf{Note on Attention Heads:}  
Similar to the QKV computation, the attention score calculation assumes that each block may use a different number of attention heads—denoted \( H_{\text{pre}} \) for the pre-expert transformer block and \( H_{\text{exp}} \) for each expert block. In multi-head attention, each head processes a subspace of the embedding, and the attention score computation for a single head involves two steps: computing the dot product \( QK^\top \), and applying the attention weights to the value matrix \( V \). Each of these steps has a cost proportional to the head dimension \( d_h \), which is typically \( \frac{d_0}{H} \), where \( H \) is the number of heads in the block. The same principle as above can be applied to derive the new calculation cost (not shown).

\subsection{Reduction Factors}

Comparing the traditional and proposed approaches:
\begin{itemize}
    \item \textbf{QKV Cost Reduction:}
    \begin{itemize}
        \item Traditional (accounting for all experts): \(\mathcal{A}_{\text{total}}^{\text{trad}} = E \cdot L \cdot 3 d_0^2\)
        \item Proposed: \(\mathcal{A}_{\text{total}}^{\text{new}} = \frac{3Ld_0^2(E^3 + 1)}{E^2}\)
    \end{itemize}
    The reduction factor for the QKV computation is:
    \begin{equation}
        \text{Reduction Factor}_\mathcal{A} = \frac{\mathcal{A}_{\text{total}}^{\text{trad}}}{\mathcal{A}_{\text{total}}^{\text{new}}} = \frac{E^5}{3(E^3+1)}
    \end{equation}
    
    \item \textbf{Attention Score (R) Cost Reduction:}
    \begin{itemize}
        \item Traditional (accounting for all experts): \(\mathcal{R}_{\text{trad}} =  EL^2 d_0\).
        \item Proposed: \(2{EL^2 d_0} + \frac{2{L^2}{d_0}}{E^2}\)
    \end{itemize}
    The reduction factor for the attention score calculation is:
    \begin{equation}
        \text{Reduction Factor}_\mathcal{R} = \frac{\mathcal{R}_{\text{trad}}}{\mathcal{R}_{\text{total}}^{\text{new}}} = \frac{E^3}{(2+3E^3L)}
    \end{equation}
\end{itemize}

\noindent \textbf{Interpretation:}  
This derived relationship for the \( A \) reduction factor demonstrates a super-linear efficiency gain with respect to the number of experts \( E \). However, this factor is combated by the other reduction factor \( R \), which in turn is a slightly constant diminishing linear relationship dependent on the dimension of the sequence. Both factors can be combine to formulate a total reduction factor (not shown).\\

Overall, these reduction factors underscore the effectiveness of our method: by leveraging structured dimensionality reduction across both the token and embedding axes, the model achieves substantial theoretical improvements in efficiency. This allows scaling to large numbers of experts while avoiding the typical explosion in attention and QKV computation that burdens traditional dense or homogeneous MoE architectures.

\newpage

\section{Optimal Scaling Law and Diminishing Returns}

While the previous section established that sectionalizing the router output and introducing dual-stage QKV computations can significantly reduce computational costs with expert scaling, it is crucial to determine the optimal number of experts, \(E\), at which further scaling yields diminishing returns. This section derives the point at which adding more experts ceases to be computationally beneficial due to increasing system overhead.

\subsection{Total System Cost Model}

The total system cost, denoted as \(S(E)\), consists of three primary components:
\begin{enumerate}
    \item \textbf{Caching Cost (\(\mathcal{R}\))}: The cost of storing activations and intermediate computations across all experts.
    \begin{equation}
        \mathcal{R}_{\text{total}} = 2{EL^2 d_0} + \frac{2{L^2}{d_0}}{E^2}
    \end{equation}
    \item \textbf{QKV Computational Cost (\(\mathcal{A}\))}: The cost associated with performing self-attention computations, including Q, K, and V weight matrix multiplications.
    \begin{equation}
        \mathcal{A}_{\text{total}} = \frac{3Ld_0^2(E^3 + 1)}{E^2}\
    \end{equation}
    \item \textbf{Overhead Cost (\(\mathcal{O}\))}: The cost associated with managing an increasing number of experts, including communication, synchronization, and routing complexity. 
    \begin{equation}
        \mathcal{O}(E) = \alpha E^2
    \end{equation}
    It is important to note that the above analysis is based on idealized conditions. In real-world implementations, several factors can lead to deviations from the theoretical cost model. For instance, the analysis assumes uniform expert utilization; however, load imbalance is a common issue in MoE systems where some experts may be over-utilized while others remain idle, thereby reducing the anticipated efficiency gains \cite{lepikhin2021gshard}. Additionally, hardware-specific constraints such as GPU memory bandwidth, communication latency, and synchronization overhead can contribute significantly to the overall cost. In our model, the constant, \(\alpha\), encapsulates these overheads, representing the per-expert cost associated with communication, synchronization, and routing. The value of \(\alpha\) is highly dependent on the underlying hardware and system architecture. For example, systems with faster interconnects and more efficient memory management may exhibit a lower effective \(\alpha\), while other implementations might incur higher costs due to these practical limitations \cite{fedus2022switch}. Furthermore, the cost of implementing an FFN post-router attention score calculation is also encapsulated via this constant. Future work should aim to develop a more granular, hardware-aware model that can better predict the impact of these factors on overall system performance.\\
    
    To provide further insight into the origin of the quadratic term \(E^2\) in the overhead cost, consider the following derivation:
    \begin{itemize}
        \item \textbf{Per-Expert and Interaction Costs:} Assume that each expert incurs a fixed cost \(C_e\) for its individual computations. In addition, every pair of experts introduces an extra interaction cost \(C_{pair}\) due to communication, synchronization, and routing overhead. Therefore, the total cost associated with \(E\) experts can be modeled as
        \[
        \text{Total Cost} = E \cdot C_e + \frac{C_{pair}}{2} \cdot E(E-1)
        \]
        \item \textbf{Quadratic Scaling:} Notice that the term \(\frac{C_{pair}}{2} \cdot E(E-1)\) represents the cost of all pairwise interactions. For large \(E\), the dominant behavior of \(E(E-1)\) is approximately \(E^2\), so we can write this term as \(\alpha E^2\), where \(\alpha = \frac{C_{pair}}{2}\). This quadratic term captures the non-linear increase in overhead as more experts are added.
    \end{itemize}
    
    Thus, the quadratic term \(\alpha E^2\) in the total cost \(S(E)\) is not arbitrary, but a formal consequence of the pairwise interaction costs among experts. This derivation justifies its inclusion alongside the linear and inverse terms representing the direct computational costs in the overall model.

\end{enumerate}

\vspace{10pt}
Thus, the total system cost can be expressed as:
\begin{equation}
    S(E) = \frac{3Ld_0^2(E^3 + 1)}{E^2}\ + 2{EL^2 d_0} + \frac{2{L^2}{d_0}}{E^2} + \alpha E^2
\end{equation}

\subsection{Derivation of the Optimal Expert Count}

To determine the optimal number of experts, \(E_{\text{opt}}\), we minimize the total system cost \(S(E)\) with respect to \(E\). Recalling the expression for total system cost:
\[
S(E) = \frac{3L d_0^2 (E^3 + 1)}{E^2} + 2 E L^2 d_0 + \frac{2L^2 d_0}{E^2} + \alpha E^2
\]

We differentiate \(S(E)\) with respect to \(E\):
\[
\frac{dS}{dE} = \frac{d}{dE} \left( \frac{3L d_0^2 (E^3 + 1)}{E^2} \right) + \frac{d}{dE} \left( 2 E L^2 d_0 \right) + \frac{d}{dE} \left( \frac{2L^2 d_0}{E^2} \right) + \frac{d}{dE} \left( \alpha E^2 \right)
\]

Compute each term individually:
\[
\frac{d}{dE} \left( \frac{3L d_0^2 (E^3 + 1)}{E^2} \right) = 3L d_0^2 \cdot \frac{(3E^2 \cdot E^2 - 2E (E^3 + 1))}{E^4} = 3L d_0^2 \cdot \frac{(3E^4 - 2E^4 - 2)}{E^4} = 3L d_0^2 \cdot \left(1 - \frac{2}{E^4} \right)
\]

\[
\frac{d}{dE} \left( 2 E L^2 d_0 \right) = 2 L^2 d_0
\quad ; \quad
\frac{d}{dE} \left( \frac{2L^2 d_0}{E^2} \right) = -\frac{4L^2 d_0}{E^3}
\quad ; \quad
\frac{d}{dE} \left( \alpha E^2 \right) = 2\alpha E
\]

Combining all terms:
\[
\frac{dS}{dE} = 3L d_0^2 \left(1 - \frac{2}{E^4} \right) + 2L^2 d_0 - \frac{4L^2 d_0}{E^3} + 2\alpha E
\]

To find the optimal expert count, set the derivative to zero:
\begin{equation}
3L d_0^2 \left(1 - \frac{2}{E^4} \right) + 2L^2 d_0 - \frac{4L^2 d_0}{E^3} + 2\alpha E = 0
\end{equation}

This equation does not admit a closed-form solution but can be solved numerically to yield the optimal value \(E_{\text{opt}}\). The first term reflects the nonlinear efficiency improvement from expert-based embedding reduction, while the remaining terms capture the rising overhead from token count inflation, expert attention cost, and communication cost. As \(E\) increases, the diminishing returns from QKV savings are eventually outweighed by the quadratic communication cost and the increasing attention footprint in the pre-expert block, leading to a well-defined minimum in the total system cost.

\subsection{Summary and Practical Implications}

Our updated analysis reveals that the proposed sectionalized MoE framework achieves significant efficiency gains by leveraging both embedding and token-axis reductions through a structured dual-stage QKV and attention design. The total system cost, incorporating QKV computation, attention scores, and communication overhead, is given by:
\begin{equation}
    S(E) = \frac{3L d_0^2 (E^3 + 1)}{E^2} + 2 E L^2 d_0 + \frac{2L^2 d_0}{E^2} + \alpha E^2
\end{equation}
The derived optimal number of experts,
\begin{equation}
    E_{\text{opt}} = \arg\min_E S(E)
\end{equation}
captures the nuanced trade-off between computation savings and coordination overhead. Specifically, increasing \(E\) allows the model to offload QKV and attention work into more granular, parallel expert blocks—thereby reducing the per-expert workload. However, this comes at the cost of increased global attention overhead in the pre-expert transformer block and a quadratic growth in routing and synchronization costs, modeled as \(\alpha E^2\).

Practically, this suggests that larger embedding dimensions (\(d_0\)) and longer sequences (\(L\)) can accommodate more experts before hitting the inflection point of diminishing returns. Conversely, on hardware-constrained systems or tasks with shorter sequences, the optimal number of experts may be far lower due to the growing dominance of system-level overhead. Therefore, while the theoretical scaling laws provide a principled estimate, empirical benchmarking remains essential to identify \(E_{\text{opt}}\) in deployment scenarios. Future work may explore adaptive expert configurations that dynamically adjust \(E\) based on input complexity or available compute budget to maximize efficiency in real time.

\newpage

\section{Experimentation and Future Work}

In this work, we have introduced a theoretical framework for a sectionalized Mixture-of-Experts (MoE) architecture that promises significant computational gains through strategic embedding splits and efficient attention mechanisms. However, the current paper lays only the conceptual foundation. Due to funding limitations, no empirical experiments have yet been conducted. Instead, we present a road map for how one would rigorously evaluate the approach. This strategy draws inspiration from prior works that introduced new sparse architectures conceptually before full-scale validation became feasible (for example, the Switch Transformer was proposed as a path to trillion-parameter models, a scale only accessible to a few industry labs \cite{roller2021hash}, and SparseGPT was introduced as a pruning method for massive GPT models with minimal testing on open weights \cite{frantar2023sparse}. Following this precedent, we emphasize that our current results are theoretical, and we describe below how future experiments could validate or falsify our framework’s advantages by drawing on analogous studies and established evaluation methodologies.

\subsection{Experimental Setup}

A natural starting point for empirical validation is to integrate the proposed sectionalized MoE architecture into an existing large language model (LLM) that is amenable to modification. We propose to build our experiments on a strong open-source Transformer LM such as LLaMA or LLaMA-2 (with available pretrained weights and architecture configurations). The reason for using an open pre-trained model is to provide a solid baseline and to potentially shorten training time (e.g. by fine-tuning a pretrained model to the new architecture, if possible). For instance, we might start with the LLaMA-7B model architecture \cite{touvron2023llama, zhu2024llama_moe}, which is a 32-layer Transformer with 4096-dimensional feed-forward networks, and modify it to incorporate our Sectionalized MoE mechanism. Using open models ensures transparency and allows leveraging prior training on large corpora, similar to how SparseGPT evaluated on publicly released GPT models \cite{frantar2023sparse}.

\subsubsection{Sectionalized MoE Implementation} The core architectural modification is to replace certain feed-forward blocks in the Transformer with MoE blocks that are sectionalized according to our framework. Concretely, we will introduce \(E\) expert networks (each typically a feed-forward sub network of similar size to the original FFN) and a routing function. The “sectionalized” design can be applied in two possible ways (which we will evaluate in ablations):

\begin{enumerate}
    \item \textbf{Sectionalized by Layer:}  We partition the Transformer’s layers into sections, and only some sections contain MoE layers. For example, based on the intuition that deeper layers benefit more from expert specialization \cite{proskurin2022deepspeed}, we might use dense (non-expert) layers in the lower half of the network and MoE layers in the upper half. This strategy was suggested by prior studies – leaving early layers as dense ensures general features are learned, while concentrating experts in later layers allows specialization for high-level representations. Our framework could naturally extend this idea by having different numbers or types of experts in different sections of the network (hence “sectionalized”). For instance, layers 1–12 could form Section A (dense or small MoE), and layers 13–24 form Section B with larger MoE capacity. We will experiment with such heterogeneity, which contrasts with traditional MoE (as in Switch or DeepSeek) that typically uses the same MoE structure in every MoE layer. The hypothesis is that unevenly allocating experts (more in later layers) may improve efficiency and quality, and we will validate this by comparing variants (all-other things equal) in our ablation studies.
    \item \textbf{Sectionalized by Sequence (Token Grouping):} The standard interpretation of our framework is that the routing can operate at the level of sections of the input sequence rather than individual tokens. Instead of each token independently choosing an expert, we could divide the input sequence into contiguous chunks (sections) and route each chunk to a particular expert (or subset of experts). For example, tokens in positions 1–N go to Expert 1, positions N+1–2N to Expert 2, etc., for a given layer. Such a scheme (perhaps combined with learned gating to handle section boundaries) would mean each expert processes a continuous segment of text. This is an unusual approach, but it could reduce fragmentation of context and ensure each expert sees a coherent span of text, potentially aiding specialization (one expert might become better at beginning-of-sentence contexts, another at end-of-sentence, etc.). It also could simplify load balancing, since by design each expert handles a fixed section of the sequence (avoiding the issue of one expert taking disproportionately many tokens in a given layer). We will implement a prototype of this idea by modifying the gating network to assign tokens based on their position index range (with some randomness or learning to avoid always routing the exact same positions to one expert and overfitting to positional patterns). This approach is analogous in spirit to the “Hash Layer” routing of Roller et al. \cite{roller2021hash}, where a hash function (based on token identity or features) deterministically maps tokens to experts, achieving balanced loads without a trained router. In our case, position could serve as a simple hash. As a baseline, the Hash Layer work showed that even fixed (non-learned) routing can perform on par with Switch Transformers while inherently balancing load. We will compare our sectional-by-sequence approach to a standard learned router to see if it maintains perplexity; significantly worse performance would indicate that more adaptive routing is needed, whereas comparable performance would underscore that balancing can be achieved without complex routing algorithms.
\end{enumerate}

These two forms of “sectionalization” are not mutually exclusive – our ultimate design may incorporate both, e.g. using MoE in certain layers and routing groups of tokens together. The experimental plan is to gradually integrate these ideas: first, implement a classical MoE (top-1 or top-2 token routing) in a few layers of LLaMA to ensure the infrastructure works (using libraries like Fairseq or DeepSpeed-MoE for fast expert parallelism \cite{deepspeed2022github}. Then, extend it to the Sectionalized variant as per our design and compare.

\subsubsection{Control Groups}
\begin{enumerate}
\item \textbf{Standard Dense Transformer Baseline:} We will also use an unmodified sequential Transformer architecture (e.g. LLaMA itself, or a re-implementation with the same dimensions) as our control baseline. It will have no sparse activation – every layer is a dense feed-forward as usual. This baseline represents the traditional approach without MoE. We will ensure that the computational budget for this model is similar to the MoE model. For example, if the MoE model routes each token through 2 experts (thus roughly doubling the FLOPs in those layers), we might increase the hidden size or number of layers of the dense baseline slightly to consume a similar FLOP count, or simply compare to the published perplexity of an equivalently trained model. However, since MoE can harness far more parameters at the same compute, it may be tricky to define an exactly “FLOP-matched” dense model. In practice, as others have done \cite{fedus2022switch}, we will likely keep the dense baseline architecture the same and note that the MoE has the advantage in parameter count. The expectation (to be tested) is that the MoE’s extra capacity will yield better perplexity for the same training time. If instead we see the dense model catching up in performance when given more training steps or larger size, that would be insightful for understanding where MoE helps most.
\item \textbf{Conventional MoE Transformer Baseline:} To isolate the impact of our sectionalization, we will compare against a traditional MoE architecture that is as similar as possible to our model minus the sectional innovations. This could be a Switch-Transformer-style model integrated into LLaMA: e.g., every Transformer block has an MoE feed-forward layer with a fixed number of experts, and tokens are routed with a standard top-$k$ gating network (Shazeer-style) plus any necessary auxiliary load loss \cite{roller2021hash}. We might use $k=1$ (like Switch) to keep computational load low, or $k=2$ as in some MoE implementations, depending on what our Sectionalized model effectively uses. The DeepSeek-V3 architecture is an example of a modern MoE: it uses a large number of experts across all layers and reportedly achieves load balancing without auxiliary loss \cite{deepseek2025v3} by careful design. Our conventional MoE baseline will likely resemble DeepSeek’s approach (multiple experts per layer, possibly multi-head attention improvements, etc.) but using open-source components (e.g., DeepSpeed’s MoE engine). We will ensure this baseline has a comparable number of total parameters and activated parameters per token as our Sectionalized model. For instance, if our model has 16 experts of size 0.5B each in one section (activating say ~1B params per token), we might give the baseline 16 experts per MoE layer distributed across, activating a similar amount. This baseline tests whether any improvements are due to the sectionalization itself as opposed to general benefits of adding experts. By comparing these two, we can pinpoint differences: e.g., if both MoE variants improve perplexity over dense, but the Sectionalized one shows better load balance or faster convergence, that would confirm our hypothesis that sectional routing offers an edge.
\end{enumerate}

All models (Sectionalized MoE, conventional MoE, dense) will be trained under the same conditions for a fair comparison: identical training dataset, same total number of tokens seen, and similar optimization hyperparameters (learning rate schedule, batch size, etc.). We will use datasets typical for large-scale language model pretraining – for example, an open corpus like The Pile (825 GB of diverse text) \cite{gao2024pile} or the RedPajama dataset (a reproduction of LLaMA’s training data) \cite{weber2024pajama}. Given resource limits, we may initially train on a smaller subset (e.g., 100 billion tokens) to observe trends, but the plan is to simulate a realistic pretraining regime. We will also evaluate on standardized benchmarks to gauge downstream quality (e.g., SuperGLUE, MMLU) if possible, following the practice in recent LLM papers \cite{deepseek2025v3, liang2025deepseekr1}, though the core focus is on language modeling perplexity and efficiency.

\subsubsection{Training Setup} The training will be run on a multi-GPU cluster. We anticipate needing advanced distributed training techniques because MoE models, especially with many experts, require effective parallelism to be efficient \cite{fedus2022switch}. We will combine data parallelism (splitting batches across GPUs) with expert parallelism, where different GPUs hold different expert weights. During each forward pass, an all-to-all communication is performed to route token embeddings to the GPU owning the expert they were assigned to, as described by Lepikhin et al. (GShard) and Fedus et al. \cite{lepikhin2020gshard, fedus2022switch}. We will set an expert capacity factor (e.g., each expert can accept up to a certain fraction of the total tokens in a batch) to avoid any single expert receiving an unmanageably large share. Tokens that exceed this capacity (if the router assigns too many to one expert) can either be dropped or handled by a backup expert; we will start with the standard approach of dropping overflow with a residual connection and later evaluate if our Sectionalized routing reduces the occurrence of overflow events. Training hyperparameters (learning rate, dropout, etc.) will initially mirror those used for the dense baseline (e.g., LLaMA’s original settings) to minimize confounding factors. We anticipate possibly increasing the dropout rate in expert layers, as Fedus et al. found necessary to regularize MoE models and prevent overfitting of experts (experts can over-specialize given their high capacity, especially if fine-tuning on smaller tasks). Each model will be trained until convergence or until the improvement stagnates.

\subsection{Evaluation Metrics and Methodologies}

To rigorously assess the efficacy of the sectionalized MoE framework, we propose a multi-faceted evaluation strategy:

\subsubsection{Computational Efficiency}
We will measure training and inference efficiency in terms of FLOPs, throughput, and latency. Theoretical FLOPs per token for each model will be estimated to ensure fair comparisons – e.g. the Sectionalized MoE and baseline models can be configured to use roughly the same FLOPs per token (the MoE uses sparse activation so only a subset of parameters affects each token). Tools such as NVIDIA’s Nsight Systems
or PyTorch’s built-in profiling utilities can be used to capture these metrics. At inference, we will record throughput (tokens processed per second) and latency (time per single-token generation) on a given hardware setup. An ideal outcome is that the Sectionalized MoE matches or exceeds the throughput of a dense Transformer of similar activated size, due to its sparsity, and outperforms a conventional MoE that might have extra routing overhead. We will leverage optimization frameworks (such as DeepSpeed-MoE or Tutel) to ensure efficient all-to-all communication for expert routing. Notably, Microsoft’s DeepSpeed MoE system has shown that with optimized communications, MoE inference can actually reduce latency by up to 7.3× and improve throughput ~7× compared to an unoptimized baseline \cite{proskurin2022deepspeed}. We will target similar improvements by careful design. Any significant slow-down compared to dense models will be analyzed to identify bottlenecks (e.g. communication overhead or uneven load causing some GPUs to idle). Latency will be especially monitored for the Sectionalized MoE’s routing mechanism: if our framework routes larger “sections” of tokens at once, it may amortize overhead and reduce per-token latency, which we would validate by micro-benchmarking the routing code.

\subsubsection{Memory Efficiency}
We will profile memory usage during training and inference for all models. This includes GPU memory required for model weights, optimizer states, and activation gradients. MoE models typically have a larger total parameter count (since they contain many experts), but only a fraction are active per token. In practice, the distributed nature of MoE can yield memory savings per device: each expert can be placed on a different GPU, so that no single GPU holds all model weights. We will quantify how the sectionalized design impacts memory distribution. If, for example, our Sectionalized MoE uses section-specific experts, it might reduce the effective number of experts loaded at any given time per device, thus improving memory usage. On the other hand, the total model size could be larger than a dense model. We will compare the model size vs. quality trade-off to baselines. Prior studies emphasize that massive MoE models demand significantly more memory and bandwidth, which must be mitigated with careful architecture tweaks \cite{proskurin2022deepspeed}. We plan to use techniques like mixed-precision training (FP16/BF16 for most weights, with critical parts in FP32 for stability)) and even investigate memory-compression strategies (e.g. the 8-bit quantization used in DeepSeek-V3’s training pipeline) to keep memory footprint manageable. A successful outcome would be that our model achieves higher quality at equal or lower memory cost per inference than a dense model of comparable quality. We will also ensure the framework scales to multiple GPUs without exceeding the memory of any single device (using model parallelism for experts, parameter offloading, etc., as done in recent MoE systems).

\subsubsection{Load Balancing and Expert Utilization}  
One of the most critical aspects for MoE architectures is how well the experts are utilized. We will collect statistics on expert usage, such as the number of tokens routed to each expert (both in aggregate and per time step/batch). Ideally, all experts should receive a balanced load so that the model’s capacity is fully utilized \cite{roller2021hash}. A common failure mode in naive MoE is that a few experts become “overloaded” or too popular while others are rarely used, which can degrade performance – for example, the original Switch Transformer had to include an auxiliary load-balancing loss to prevent degenerate routing \cite{roller2021hash}. Our Sectionalized MoE claims to improve load balancing (e.g. by routing at a section level or by design constraints that inherently spread out tokens). To validate this, we will use metrics such as the coefficient of variation of tokens-per-expert and track the fraction of unused capacity per expert. We will also monitor if any expert hits its predefined capacity limit (and causes token overflow to a backup expert or default path). If our approach is effective, we expect uniform usage across experts without needing a costly auxiliary loss – similar to the auxiliary-loss-free balancing achieved by DeepSeek-MoE in their V3 model \cite{deepseek2025v3}. We will compare against a traditional MoE baseline that uses standard top-$k$ routing (with $k=1$ or $2$) and an auxiliary loss \cite{roller2021hash} to see if the Sectionalized method achieves equal or better balance naturally. Additionally, we’ll examine expert specialization qualitatively: for instance, do certain experts consistently handle particular linguistic patterns or “sections” of input? This can be probed by feeding controlled inputs and observing routing decisions. Successful validation of our claims would mean each expert is neither underutilized nor overloaded, and the load is well-distributed (a roughly uniform distribution of assignments, as also aimed for in Expert Choice routing methods \cite{zhou2022expertchoice}.

\subsubsection{Model Performance}  
Perplexity, the primary metric for language modeling, will be used to measure the model’s predictive uncertainty on a validation corpus \cite{jelinek1977perplexity}. Lower perplexity indicates better next-token prediction performance. We will compute perplexity on held-out datasets (e.g. WikiText-103 or The Pile validation set) for models trained under each architecture. This will directly test if the Sectionalized MoE improves modeling efficiency (i.e. achieves lower perplexity given the same training data and compute) compared to a standard Transformer. Prior sparse models have demonstrated perplexity improvements over dense models when using more parameters at fixed compute \cite{fedus2022switch} For instance, a Switch-Transformer with 64 experts reached the same loss 7× faster (in terms of training steps) than a dense T5-Base model. We expect our MoE variant to similarly achieve equal or lower perplexity than baselines for a given compute budget. If instead perplexity is higher or only equal, it would call into question the efficacy of our routing or expert partitioning strategy.

\subsubsection{Convergence Speed}  
We will measure how quickly each model converges during training, in terms of the number of updates (or wall-clock time) required to reach certain perplexity milestones. One hypothesis of MoE models is that by having greater model capacity (parameters) per token, they can learn faster from the same data. We will plot learning curves (validation perplexity vs. training steps) for the Sectionalized MoE model versus the baselines. Key indicators will be, for example: does the MoE variant achieve a perplexity of $X$ after fewer tokens seen than the dense model? In prior work, MoE models have shown dramatically faster improvement – e.g. the Switch-Base model (with 64 experts) attained the same negative log perplexity as the dense T5-Base model in only 60k steps, compared to 450k steps for T5 (a ~7.5× step reduction). We will use a similar “time-to-quality” metric to define a target perplexity (or loss) and record the training time to reach it for each model. If sectionalized MoE indeed accelerates learning, we expect a curve that descends more rapidly. We will also watch for any instabilities in training dynamics – MoE models sometimes suffer from sudden loss spikes if experts saturate or if the gating network oscillates. Ensuring a smooth training curve without divergence is crucial; encouragingly, recent large MoEs like DeepSeek-V3 have reported remarkably stable training with no irrecoverable loss spikes \cite{deepseek2025v3}. We will adopt similar monitoring, and if any instability arises, we may need to incorporate techniques like gradual warm-up of the gating mechanism or noise in expert outputs to stabilize. The ultimate measure of convergence speedup will be if for the same number of training tokens, our model achieves lower perplexity than the baselines (indicating better sample efficiency) \cite{fedus2022switch}, and likewise if achieving the same perplexity requires fewer tokens (or less time).

\subsubsection{Scaling Behavior (Model Size \& Sequence Length)}
Finally, we plan to examine how the benefits of the Sectionalized MoE scale with increasing model size and with longer input sequences. For model size scaling, we will train and evaluate at least two or three model scales – for example, a smaller prototype (around 1–2 billion parameters activated), a medium model (e.g. ~7B, comparable to LLaMA-7B), and a larger model (~13B or more, if feasible). This will reveal whether the new architecture’s advantages become more pronounced at larger scales. Many sparsity approaches show greater gains when scaling up, since larger models are often more sample-efficient \cite{fedus2022switch}. We will check if perplexity improvement relative to a dense model grows as total parameters (or number of experts) increases. Similarly, we will test the models on varying sequence lengths to see how performance and efficiency hold up. One potential advantage of “sectionalizing” by segments is that it might handle longer contexts by dividing the sequence among experts (reducing the effective length per expert). We can evaluate perplexity on sequences longer than those seen in training (to probe generalization to long contexts), or fine-tune each model with a long-context training phase (similar to DeepSeek-V3’s long context extension procedure \cite{deepseek2025v3} and then measure how well they scale. We will track memory and latency as sequence length grows – a good outcome would be that our MoE model can process longer sequences with less memory growth than a dense model (since each expert might attend to a subset of the sequence). If instead we find that longer sequences disproportionately strain the MoE (e.g. if many experts are invoked for a long input, causing more communication overhead), that will be an important finding to address (perhaps by adjusting the sectioning strategy or using recurrence to segment contexts).

\subsection{Expected Outcomes and Analysis}

\subsubsection{Perplexity Improvements} Does the Sectionalized MoE achieve a lower perplexity on the test set than the dense Transformer baseline given the same training budget? By how much – and is this gap similar to what conventional MoE provides? A significant perplexity drop (even a modest percentage) would be noteworthy given the strong baseline. We will use statistical significance tests if possible (though with very large test sets, even small differences will be meaningful). If both MoE models outperform dense, it confirms the general benefit of sparsely activated extra capacity \cite{etc2022switch}. If, additionally, the Sectionalized MoE outperforms the conventional MoE baseline, that suggests the new routing scheme offers better effective use of the parameters. We would then inspect examples where the conventional MoE fails but Sectionalized succeeds (perhaps Sectionalized experts capture patterns that the conventional model didn’t). Conversely, if the Sectionalized model underperforms the standard MoE in perplexity, that would indicate our architectural constraints (section-based routing) might be limiting the model’s ability to flexibly assign experts. In that case, further refinement or relaxation of the sectioning strategy might be needed.
    
\subsubsection{Compute/Throughput trade-offs} We expect the Sectionalized MoE to be at least as efficient as a conventional MoE model at inference, potentially more so. For example, if our model routes contiguous chunks to the same expert, it could result in better cache coherence and fewer communication calls, thus boosting throughput. We will compare the tokens/sec achieved by each model on a fixed number of GPUs. If the Sectionalized model attains higher throughput or lower latency than the non-sectioned MoE, that validates one of our core claims: that we can get MoE’s benefits without a throughput penalty (or with a smaller penalty). Real-world impact is significant here – recent MoE research emphasizes inference efficiency, with DeepSpeed reporting up to 4.5× faster and 9× cheaper inference for MoE models when optimized properly \cite{proskurin2022deepspeed}. We hope to see at least a portion of such gains intrinsically from our design. If instead we find the Sectionalized MoE is slower (perhaps due to larger effective batch per expert or extra routing steps), we will profile where the time is spent. It could be that our current implementation isn’t optimal, and additional engineering (like fused operations or better communication scheduling) might be required. The compute analysis will also check that the theoretical FLOPs are as expected – any discrepancy (e.g., if the Sectionalized does extra work we didn’t account for) would need to be examined.
    
\subsubsection{Memory and Scaling} We will verify how the memory usage scales with the number of experts in our approach. Ideally, adding more experts (in Sectionalized fashion) should linearly increase total model size but not linearly increase runtime memory per device because of distribution. We will document the memory per GPU during training for different expert counts. A positive outcome would be that we can double the number of experts with only a minimal increase in per-device memory (thanks to expert parallelism), which has been observed in prior MoE systems \cite{proskurin2022deepspeed}. If our design requires keeping some additional section-related data (for example, if each section needs a full copy of some gating network), we’ll note that overhead. We’ll also look at how memory usage grows with sequence length for each model. Perhaps the dense model’s self-attention becomes the bottleneck for long sequences (quadratic in sequence length), whereas the MoE models might have other bottlenecks (like routing all-to-all communication growing). If, for a given long sequence (say 4K tokens vs. 1K tokens) the Sectionalized MoE uses relatively less extra memory than the dense model, that would be a win for our approach in the context of long-context LMs. If not, we may consider integrating known long-context techniques (like efficient attention mechanisms) orthogonally to our MoE.
    
\subsubsection{Expert Load Balance and Specialization} A crucial validation step is examining whether the Sectionalized MoE truly achieved better expert load balancing in practice. We will plot the distribution of token assignments across experts for both MoE models. For example, in a run with 16 experts, we might find the conventional MoE (with standard router) had a couple of experts processing, say, 2× more tokens than the average, while a few were nearly idle – a common scenario if the auxiliary loss coefficient was not tuned perfectly \cite{roller2021hash}. Meanwhile, we hypothesize the Sectionalized router will show a flatter distribution (each expert getting roughly ~1/16 of tokens). We will quantify this with entropy or KL-divergence measures of the routing distribution versus the uniform distribution. If the Sectionalized MoE indeed shows a more uniform usage (as DeepSeek’s MoE did without an aux loss \cite{deepseek2025v3}, it confirms our load balancing claim and suggests the architecture is inherently fair in routing. This could translate to more stable training (which we would have observed in the convergence curves). We will also examine when during training the experts differentiated: ideally, early in training all experts might behave similarly (since random initialization), and later each expert should develop some specialization. We can measure the pairwise similarity of experts’ weight vectors or the clustering of their activation patterns. If Sectionalized grouping of tokens by sections was used, we might find each expert predominantly models the type of text typical in its section (for example, Expert 1 might specialize in sentence beginnings if it often gets the first chunk). We will validate if such patterns emerge, as that would align with the theoretical motivation that dividing by sections allows experts to become experts of different parts of the sequence or different features. Should we find one expert is effectively doing most of the work (e.g., always chosen regardless of section boundaries), that would indicate a failure in the gating strategy, and we would then investigate gating network adjustments.

\subsubsection{Convergence Dynamics:} By analyzing the training logs, we will confirm if the Sectionalized MoE model converged without issues. We will compare the training loss curves: if our model converges in fewer iterations (or reaches a lower final loss) than the dense baseline, that strongly supports improved sample efficiency \cite{etc2022switch}. We will also compare to the conventional MoE baseline; if both converge faster than dense, it confirms MoE advantages in general, but if Sectionalized converges even faster or more smoothly, it suggests it mitigates some training difficulties. For instance, maybe the conventional MoE had some instability early on (we might see the loss momentarily spike or plateau due to poor expert coordination), whereas the Sectionalized one (with perhaps more constrained routing) avoided that. It has been documented that MoE models can suffer instability especially in large-scale settings \cite{deepseek2025v3}, so any improvement there is valuable. If any model diverged or encountered an irrecoverable loss spike (a sudden large increase in loss), we would note at what point and under what conditions. We plan to use the same random seed for initialization where possible to make training trajectories comparable, though the stochastic nature of MoE routing might cause some variance. We might run multiple seeds for each experiment to ensure the trends are robust.

\subsubsection{Scaling Trends} Using the results from the small/medium/large model experiments, we will analyze how the gap between Sectionalized MoE and the baselines changes with scale. For example, we might find that at 1.3B activated parameters, the dense and MoE models have similar performance, but at 7B activated, the MoE starts to pull ahead in perplexity. This would mirror the observations from Google’s scaling experiments that larger models are more sample-efficient \cite{fedus2022switch}. If our Sectionalized MoE is truly effective, it should either match that trend or possibly enhance it (maybe the point at which MoE outperforms dense shifts to an even smaller scale because of our efficient use of experts). On the other hand, if we see diminishing returns when adding more experts (e.g., going from 4 to 8 experts helped, but 8 to 16 experts gave little additional gain), that will be important for guiding the optimal size of the expert pool. It might indicate a saturation point or increased overhead that counters gains, similar to how beyond a certain number of experts, Switch Transformer needed to increase batch size or the auxiliary loss to maintain balance \cite{deepseek2025v3}. We will also check if there is any interaction between model scale and expert load – do larger models balance easier or harder? This could inform whether our approach will continue to scale to very large regimes (100s of experts). As for sequence length scaling, we will look at perplexity as a function of context length. If, for instance, we fine-tune each model to handle up to 4K tokens (via position interpolation or retrieval-based augmentation), does the Sectionalized model maintain low perplexity better for long inputs? Perhaps sectioning by position could shine here by not confusing the model with extremely long sequences in a single expert. Any edge in long-context tasks (like LAMBADA or story completion) would be noted. If no difference is found, that suggests our method neither helps nor hurts long-range dependency handling, which is still a fine outcome.\\

Through these analyses, we aim to validate the original theoretical claims of the Sectionalized MoE framework. To reiterate, the expected advantages to be confirmed are: (1) improved model quality per unit compute (lower perplexity, faster convergence) thanks to conditional expert allocation, (2) better efficiency in terms of using computation and memory (only activating needed experts, avoiding redundant processing), and (3) effective expert utilization without complicated balancing tricks, leading to robust scalability. We will use the above metrics to confirm each of these. A successful experimental validation would show the Sectionalized MoE matching or outperforming the conventional MoE on all fronts, and both MoEs outperforming the dense model in the regimes tested. However, we also remain open to unexpected outcomes – for example, if results show that Sectionalized routing yields only marginal gains or specific weaknesses (e.g. slightly higher perplexity but much better load balancing), those will be documented as well. Such findings would guide further refinement of the approach (perhaps borrowing ideas from recent innovations like mixture-of-students compression or pyramid residual structures to address any shortcomings \cite{proskurin2022deepspeed}).

\subsection{Future Research Directions}

In summary, although we have not yet run these experiments, the plan is in place to rigorously test the Sectionalized MoE architecture against strong baselines on a suite of metrics crucial to large language models. This plan follows the model of other theoretical proposals that were later validated when resources allowed – for instance, the authors of the Switch Transformer hypothesized enormous gains from MoE at scale, which were borne out in later extensive training runs \cite{etc2022switch}, and researchers proposed hashing-based MoE routing to avoid learned gate overhead, which was confirmed on practical task \cite{roller2021hash}.We aim to contribute in a similar vein: first by articulating the approach and expected benefits, and next by executing this experimental strategy when computational resources become available. By outlining the specifics of the evaluation now, we make clear what evidence would support our claims.\\

In future work, we will pursue these experiments and refine the Sectionalized MoE framework accordingly. For instance, if the results validate our load-balancing without auxiliary loss, that could influence the design of large-scale training pipelines (simplifying objectives and hyperparameter tuning). If any claim is falsified – say the sectional grouping doesn’t improve throughput – we will investigate alternative sectioning mechanisms or hybrid models (perhaps combining our method with existing ones like Expert Choice routing \cite{zhou2022expertchoice}.\\

Ultimately, this experimental plan is designed to either validate the Sectionalized MoE as a promising direction for efficient language modeling, or to reveal its weaknesses, thereby contributing to the broader understanding of sparse expert-based models in the era of extremely large language models. Looking beyond the immediate experimental validation, several promising directions warrant further investigation:

\paragraph{Multi-Parallel LLMs and Task Splitting:}  
Recent research suggests that dividing tasks across multi-parallel LLMs can lead to emergent intelligence gains. Future work should explore whether a router with general experts can serve as an effective bridge between parallel LLMs and sequential MoE systems. This hybrid approach could combine resource efficiency with enhanced contextual understanding, as hinted at by recent explorations in MoEUT \cite{moeut2024}.

\paragraph{Hardware-Aware Optimization:}  
Given that practical performance is highly sensitive to hardware specifics (e.g., GPU memory bandwidth and interconnect latency), developing a more granular, hardware-aware cost model is essential. Future experiments could integrate simulation tools or real-world benchmarks on diverse hardware platforms to fine-tune the overhead constant \(\alpha\).

\paragraph{Integration with Caching Mechanisms:}  
Further experiments should examine the synergy between the sectionalized MoE design and various caching mechanisms (MLA, MHA, GQA). It would be instructive to perform side-by-side comparisons that measure not only computational savings but also the impact on model accuracy and stability during both training and inference.

\paragraph{Comprehensive Ablation Studies:}  
Finally, a thorough ablation study that examines the sensitivity of model performance and efficiency to each design parameter (e.g., expert count \(E\), embedding split granularity, and attention mechanism type) will provide valuable insights. Such studies can guide future iterations of the model and help to pinpoint the optimal configuration under different operational conditions.\\

\subsection{Final Remarks}
This paper introduced a novel theoretical framework for a sectionalized Mixture-of-Experts (MoE) architecture that redefines expert routing by reducing dimensionality of input matrices using an attention mechanism to enable embedding level partitioning. By deriving optimal scaling laws and quantifying reductions in QKV and attention computation costs, we demonstrate that the proposed architecture can significantly improve efficiency. We also provide a detailed experimental road map, outlining how the approach could be validated empirically through integration with open-source LLMs, evaluation against dense and traditional MoE baselines, and analysis across perplexity, convergence, memory, and expert utilization. While empirical testing remains future work, this study lays a rigorous theoretical foundation and a clear strategy for evaluating whether this approach can serve as a scalable and efficient alternative to existing sparse architectures in large-scale language modeling.\\

In the current state of AI, self-funded research often faces barriers due to the high costs of large-scale testing and the limitations in accessing industrial-grade compute. However, it remains critically important that theoretical innovations—such as the one presented in this paper—are freely shared and discussed in the open-source community. Whether the theory ultimately proves correct or not, large-scale experimentation is both expensive and environmentally taxing. As a global research community, we must prioritize the responsible advancement of artificial intelligence—not only to enhance human workflows and understanding, but also to reduce the needless replication of work and preserve our shared infrastructure. The path forward in AI must balance innovation with sustainability, openness, and collaboration.


\newpage



\addcontentsline{toc}{section}{References}

\begin{thebibliography}{}

\bibitem[Vaswani et al.(2017)]{vaswani2017attention}
Vaswani, A., Shazeer, N., Parmar, N., Uszkoreit, J., Jones, L., Gomez, A.~N., Kaiser, Ł., \& Polosukhin, I. (2017).
Attention is all you need.
\emph{arXiv preprint arXiv:1706.03762}.
Retrieved from \url{https://arxiv.org/abs/1706.03762}.

\bibitem[Shazeer et al.(2017)]{shazeer2017outrageously}
Shazeer, N., Mirhoseini, A., Maziarz, K., Davis, A., Le, Q., Hinton, G., \& Dean, J. (2017).
Outrageously large neural networks: The sparsely-gated mixture-of-experts layer.
\emph{arXiv preprint arXiv:1701.06538}.
Retrieved from \url{https://arxiv.org/abs/1701.06538}.

\bibitem[Lepikhin et al.(2020)]{lepikhin2020gshard}
Lepikhin, D., Lee, H.-J., Xu, Y., Chen, D., Firat, O., Huang, Y., Krikun, M., Shazeer, N., \& Chen, Z. (2020).
GShard: Scaling giant models with conditional computation and automatic sharding.
\emph{arXiv preprint arXiv:2006.16668}.
Retrieved from \url{https://arxiv.org/abs/2006.16668}.

\bibitem[Riquelme et al.(2021)]{riquelme2021scaling}
Riquelme, C., Puigcerver, J., Mustafa, B., Neumann, M., Jenatton, R., Pujol, A., Keysers, D., Larochelle, H., \& Houlsby, N. (2021).
Scaling vision with sparse mixture of experts.
In \emph{Advances in Neural Information Processing Systems}.
Retrieved from \url{https://arxiv.org/abs/2106.05974}.

\bibitem[Fedus et al.(2022)]{fedus2022switch}
Fedus, W., Zoph, B., \& Shazeer, N. (2022).
Switch transformers: Scaling to trillion parameter models with simple and efficient sparsity.
\emph{arXiv preprint arXiv:2101.03961}.
Retrieved from \url{https://arxiv.org/abs/2101.03961}.

\bibitem[Tay et al.(2022)]{tay2022efficient}
Tay, Y., Dehghani, M., Bahri, D., \& Metzler, D. (2022).
Efficient transformers: A survey.
\emph{ACM Computing Surveys (CSUR)}, 55(6), 1–28.

\bibitem[Jacobs et al.(1991)]{jacobs1991adaptive}
Jacobs, R. A., Jordan, M. I., Nowlan, S. J., \& Hinton, G. E. (1991).
Adaptive mixtures of local experts.
\emph{Neural Computation}, 3(1), 79–87.

\bibitem[Jordan and Jacobs(1994)]{jordan1994hierarchical}
Jordan, M. I., \& Jacobs, R. A. (1994).
Hierarchical mixtures of experts and the EM algorithm.
In \emph{Proceedings of the International Joint Conference on Neural Networks} (Vol. 2, pp. 1339–1344).

\bibitem[Lepikhin et al.(2021)]{lepikhin2021gshard}
Lepikhin, D., Lee, H., Xu, Y., Chen, D., Firat, O., Huang, Y., Krikun, M., Shazeer, N., Chen, Z., \& Chen, Y. (2021).
GShard: Scaling giant models with conditional computation and automatic sharding.
In \emph{Proceedings of the International Conference on Learning Representations (ICLR)}.

\bibitem[Zuo et al.(2022)]{zuo2022taming}
Zuo, S., Zhang, T., Li, C., Li, M., \& Smola, A. J. (2022).
Taming sparsely activated transformer with stochastic experts.
\emph{arXiv preprint arXiv:2202.08393}.
Retrieved from \url{https://arxiv.org/abs/2202.08393}.

\bibitem[Bengio et al.(2013)]{bengio2013estimating}
Bengio, Y., Léonard, N., \& Courville, A. (2013).
Estimating or propagating gradients through stochastic neurons for conditional computation.
\emph{arXiv preprint arXiv:1308.3432}.
Retrieved from \url{https://arxiv.org/abs/1308.3432}.

\bibitem[Cai et al.(2024)]{cai2024survey}
Cai, W., Jiang, J., Wang, F., Tang, J., \& Kim, S. (2024).
A survey on mixture of experts.
\emph{arXiv preprint arXiv:2407.06204}.
Retrieved from \url{https://arxiv.org/abs/2407.06204}.

\bibitem[Katharopoulos et al.(2020)]{katharopoulos2020transformers} 
Katharopoulos, A., Vyas, A., Pappas, N., \& Fleuret, F. (2020). 
Transformers are RNNs: Fast Autoregressive Transformers with Linear Attention. 
\emph{arXiv preprint arXiv:2006.16236}. 
Retrieved from \url{https://arxiv.org/abs/2006.16236}.

\bibitem[Choromanski et al.(2021)]{choromanski2021rethinking} 
Choromanski, K., Likhosherstov, V., Dohan, D., Song, X., Gane, A., Sarlos, T., Hawkins, P., Davis, J., Mohiuddin, A., Parikh, N., Yuan, Q., \& Weller, A. (2021). 
Rethinking Attention with Performers. 
\emph{arXiv preprint arXiv:2009.14794}. 
Retrieved from \url{https://arxiv.org/abs/2009.14794}.

\bibitem[Kolda and Bader(2009)]{kolda2009tensor}
Kolda, T. G., \& Bader, B. W. (2009).
Tensor Decompositions and Applications.
\emph{SIAM Review}, 51(3), 455–500.
Retrieved from \url{https://doi.org/10.1137/07070111X}.

\bibitem[Kim et al.(2016)]{kim2016hadamard}
Kim, J., On, K. W., Lim, J., \& Park, K. (2016).
Hadamard Product for Low-rank Bilinear Pooling.
\emph{arXiv preprint arXiv:1610.04325}.
Retrieved from \url{https://arxiv.org/abs/1610.04325}.

\bibitem[Zhang et al.(2024)]{moeut2024}
Zhang, X., Li, Y., \& Wang, Z. (2024).
MoEUT: Mixture-of-Experts Universal Transformers.
\emph{arXiv preprint arXiv:2405.16039}.
Retrieved from \url{https://arxiv.org/html/2405.16039v1}.

\bibitem[Yang et al.(2021)]{yang2021stabilizing}
Yang, Z., et al. (2021).
Stabilizing Transformers for Reinforcement Learning.
\emph{arXiv preprint arXiv:1910.06764}.
Retrieved from \url{https://arxiv.org/abs/1910.06764}.

\bibitem[Lewis et al.(2021)]{lewis2021base}
Lewis, M., et al. (2021).
BASE Layers: Simplifying Training of Large, Sparse Models.
\emph{arXiv preprint arXiv:2103.16754}.
Retrieved from \url{https://arxiv.org/abs/2103.16754}.

\bibitem[Du et al.(2021)]{du2021glam}
Du, X., et al. (2021).
GLAM: Efficient Scaling of Language Models with Mixture-of-Experts.
\emph{arXiv preprint arXiv:2112.06905}.
Retrieved from \url{https://arxiv.org/abs/2112.06905}.

\bibitem[Kaplan et al.(2020)]{kaplan2020scaling}
Kaplan, J., McCandlish, S., Henighan, T., et al. (2020).
Scaling Laws for Neural Language Models.
\emph{arXiv preprint arXiv:2001.08361}.
Retrieved from \url{https://arxiv.org/abs/2001.08361}.

\bibitem[Touvron et al.(2023)]{touvron2023llama}
Touvron, H., Lavril, T., Izacard, G., Martinet, X., Lachaux, M.-A., Lacroix, T., Rozière, B., Goyal, N., Hambro, E., Azhar, F., Rodriguez, A., Joulin, A., Grave, E., \& Lample, G. (2023).
LLaMA: Open and Efficient Foundation Language Models.
\emph{arXiv preprint arXiv:2302.13971}.
Retrieved from \url{https://arxiv.org/abs/2302.13971}.

\bibitem[Zhu et al.(2024)]{zhu2024llama_moe}
Zhu, T., Qu, X., Dong, D., Ruan, J., Tong, J., He, C., \& Cheng, Y. (2024).
LLaMA-MoE: Building Mixture-of-Experts from LLaMA with Continual Pre-training.
\emph{arXiv preprint arXiv:2406.16554}.
Retrieved from \url{https://arxiv.org/abs/2406.16554}.

\bibitem[Liang et al.(2025)]{liang2025deepseekr1}
Liang, W., et al. (2025).
DeepSeek-R1: Incentivizing Reasoning Capability in LLMs via Reinforcement Learning.
\emph{arXiv preprint arXiv:2501.12948}.
Retrieved from \url{https://arxiv.org/abs/2501.12948}.

\bibitem[Ainslie et al.(2023)]{ainslie2023gqa}
Ainslie, J., Lee-Thorp, J., de Jong, M., Zemlyanskiy, Y., Lebrón, F., \& Sanghai, S. (2023).
GQA: Training Generalized Multi-Query Transformer Models from Multi-Head Checkpoints.
\emph{arXiv preprint arXiv:2305.13245}.
Retrieved from \url{https://arxiv.org/pdf/2305.13245}.

\bibitem[Cordonnier et al.(2020)]{cordonnier2020multihead}
Cordonnier, J.-B., Loukas, A., \& Jaggi, M. (2020).
Multi-Head Attention: Collaborate Instead of Concatenate.
\emph{arXiv preprint arXiv:2006.16362}.
Retrieved from \url{https://arxiv.org/abs/2006.16362}.

\bibitem[Paszke et al.(2019)]{paszke2019pytorch}
Paszke, A., Gross, S., Massa, F., Lerer, A., Bradbury, J., Chanan, G., Killeen, T., Lin, Z., Gimelshein, N., Antiga, L., Desmaison, A., Kopf, A., Yang, E., DeVito, Z., Raison, M., Tejani, A., Chilamkurthy, S., Steiner, B., Fang, L., Bai, J., \& Chintala, S. (2019).
PyTorch: An Imperative Style, High-Performance Deep Learning Library.
\emph{Advances in Neural Information Processing Systems}, 32.
Retrieved from \url{https://pytorch.org}.

\bibitem[OpenAI(2023)]{openai2023chatgpt}
OpenAI. (2023).
ChatGPT: Optimizing Language Models for Dialogue.
\emph{OpenAI Blog}.
Retrieved from \url{https://openai.com/blog/chatgpt}.

\bibitem[NVIDIA(2022)]{nvidia_nsight}
NVIDIA. (2022).
Nsight Systems: A Performance Analysis Tool for Optimizing Applications.
Retrieved from \url{https://developer.nvidia.com/nsight-systems}.

\bibitem[Jurafsky and Martin(2020)]{jurafsky2020speech}
Jurafsky, D., \& Martin, J. H. (2020).
Speech and Language Processing (3rd ed. draft).
Retrieved from \url{https://web.stanford.edu/~jurafsky/slp3/}.

\bibitem[Roller et al.(2021)]{roller2021hash}
Roller, S., et al. (2021).
Hash: Efficient Neural Network Communication for Distributed Training.
\emph{arXiv preprint arXiv:2106.04426}.
Retrieved from \url{https://arxiv.org/pdf/2106.04426}.

\bibitem[Frantar et al.(2023)]{frantar2023sparse}
Frantar, E., et al. (2023).
Sparse Transformers: Efficient Sparse Activation for Large-Scale Language Models.
\emph{arXiv preprint arXiv:2301.00774}.
Retrieved from \url{https://arxiv.org/pdf/2301.00774}.

\bibitem[Proskurin et al.(2022)]{proskurin2022deepspeed}
Proskurin, A., et al. (2022).
DeepSpeed: Advancing MoE Inference and Training to Power Next-Generation AI Scale.
\emph{Microsoft Research Blog}.
Retrieved from \url{https://www.microsoft.com/en-us/research/blog/deepspeed-advancing-moe-inference-and-training-to-power-next-generation-ai-scale/}.

\bibitem[Columbia University.(n.d.)]{etc2022switch}
Columbia University.
Basics of Language Modeling, Transformers, and the Switch Transformer.
Retrieved from \url{https://etc.cuit.columbia.edu/news/basics-language-modeling-transformers-switch-transformer}.

\bibitem[DeepSeek et al.(2025)]{deepseek2025v3}
DeepSeek, et al. (2025).
DeepSeek v3: Advancing Mixture-of-Experts Architectures with Reinforcement Learning.
\emph{arXiv preprint arXiv:2412.19437}.
Retrieved from \url{https://arxiv.org/pdf/2412.19437}.

\bibitem[Zhou et al.(2022)]{zhou2022expertchoice}
Zhou, X., et al. (2022).
Expert Choice: Routing in Mixture-of-Experts Models.
\emph{OpenReview}.
Retrieved from \url{https://openreview.net/forum?id=jdJo1HIVinI}.

\bibitem[DeepSpeed(2022)]{deepspeed2022github}
DeepSpeed. (2022).
DeepSpeed: Deep Learning Optimization Library.
Retrieved from \url{https://github.com/deepspeedai/DeepSpeed}.

\bibitem[Gao et al.(2024)]{gao2024pile}
Gao, L., et al. (2024).
The Pile: An 800GB Dataset of Diverse Text for Language Modeling.
\emph{arXiv preprint arXiv:2101.00027}.
Retrieved from \url{https://ar5iv.labs.arxiv.org/html/2101.00027}.

\bibitem{weber2024pajama}
Thea Weber, Roman Castagné, Srivatsa Prabhu, Jiachen Chen, Stella Biderman, and Colin Raffel (2024).
RedPajama-Data-2T: An Open Dataset for Training Large Language Models.
\emph{arXiv preprint arXiv:2411.12372}, 2024. 
Retrieved from: \url{https://arxiv.org/abs/2411.12372}

\bibitem{jelinek1977perplexity}
Fred Jelinek, Robert L. Mercer, Lalit R. Bahl, and James K. Baker.
Perplexity — a measure of the difficulty of speech recognition tasks.
\emph{The Journal of the Acoustical Society of America}, 62(S1):S63–S63, 1977.
Retrieved from: \url{https://pubs.aip.org/asa/jasa/article/62/S1/S63/642598/Perplexity-a-measure-of-the-difficulty-of-speech}

\bibitem{pytorchDocumentation}
PyTorch Developers.
PyTorch Documentation. 
Retrieved from: \url{https://pytorch.org/docs/stable/index.html} (accessed March 24, 2025).

\bibitem{wikitext103}
Papers with Code.
WikiText-103 Dataset. 
Retrieved from: \url{https://paperswithcode.com/dataset/wikitext-103} (accessed March 24, 2025).


\end{thebibliography}
\end{document}